\documentclass[journal]{IEEEtran}

\pdfoutput=1
\usepackage{slashbox}
\usepackage[all]{xy}
\usepackage{setspace}
\usepackage{amssymb}
\usepackage{color}
\usepackage{algorithm}
\usepackage{algorithmic}
\usepackage{cite}
\usepackage{hyperref}
\usepackage{cleveref}
\usepackage{graphicx}  

\usepackage{subfigure} 

\usepackage{amsmath}   

\begin{document}
%
\title{Target Oriented High Resolution SAR Image Formation via Semantic Information Guided Regularizations}

\author{Biao Hou,~\IEEEmembership{Member,~IEEE,}
        Zaidao Wen,
        Licheng Jiao,~\IEEEmembership{Senior Member,~IEEE,}
        and~Qian Wu
}
\maketitle

\begin{abstract}
Sparsity-regularized synthetic aperture radar (SAR) imaging framework has shown its remarkable performance to generate a feature enhanced high resolution image, in which a sparsity-inducing regularizer is involved by exploiting the sparsity priors of some visual features in the underlying image. However, since the simple prior of low level features are insufficient to describe different semantic contents in the image, this type of regularizer will be incapable of distinguishing between the target of interest and unconcerned background clutters. As a consequence, the features belonging to the target and clutters are simultaneously affected in the generated image without concerning their underlying semantic labels. To address this problem, we propose a novel semantic information guided framework for target oriented SAR image formation, which aims at enhancing the interested target scatters while suppressing the background clutters. Firstly, we develop a new semantics-specific regularizer for image formation by exploiting the statistical properties of different semantic categories in a target scene SAR image. In order to infer the semantic label for each pixel in an unsupervised way, we moreover induce a novel high-level prior-driven regularizer and some semantic causal rules from the prior knowledge. Finally, our regularized framework for image formation is further derived as a simple iteratively reweighted $\ell_1$ minimization problem which can be conveniently solved by many off-the-shelf solvers. Experimental results demonstrate the effectiveness and superiority of our framework for SAR image formation in terms of target enhancement and clutters suppression, compared with the state of the arts. Additionally, the proposed framework opens a new direction of devoting some machine learning strategies to image formation, which can benefit the subsequent decision making tasks.
\end{abstract}

\IEEEpeerreviewmaketitle

\section{Introduction}\label{sec:introduction}
\PARstart{A}{utomatic} target recognition (ATR) is one of the most {important} decision making tasks for synthetic aperture radar (SAR), in which a high quality SAR image is required to provide some informative target features for recognition \cite{Hummel2000}. Therefore, the SAR platform operating in the spotlight mode is widely leveraged for ATR \cite{Hummel2000} since it can generate a target image with higher resolution by continuously illuminating the target scene from a series of viewing angles \cite{Carrara1995}. Conventionally, forming a SAR image, {namely SAR imaging,} relies on inverse Fourier transformation for the spotlight mode SAR, e.g., polar formatting algorithm and convolution back-projection algorithm \cite{Carrara1995,Desai1992Convolution}. These approaches have been extensively leveraged for SAR image formation due to their simplicity and efficiency, but they still suffer from the following deficiencies in terms of the ATR task. 1). They rely on a perfectly received and sampled data to form a high quality image, which makes them sensitive to various non ideal or noisy environments such as limitations in the sampled data as well as viewing angles. In these scenarios, the quality and resolution of the obtained image will be generally degraded. 2). {The contents of the underlying scene image and the features of the target are not concerned in these imaging algorithms so that the generated image will not provide a positive contribution to improve the performance of ATR or other decision-making tasks, e.g. segmentation etc.}
\par These deficiencies consequently raise an issue whether we could develop a SAR imaging framework that is driven by the following decision-making tasks \cite{Cetin2014}. More specifically, for example, most ATR algorithms will generally exploit some features of the target extracted from a high resolution SAR image, such as the scattering points configuration, target contour and shape \cite{Hummel2000,Zhang2014Attributed}. If the imaging algorithm can take these features into consideration and simultaneously provide a feature enhanced target image, the subsequent ATR will be easier. For this purpose, $\c{C}$etin and Karl \cite{cetin2001feature} propose a promising feature enhanced SAR imaging framework which recasts the imaging procedure as solving a regularized linear inverse problem. This framework enables us to enhance some task-specific features via involving a variety of regularization functions \cite{franklin1974tikhonov}. In their framework, they adopt the $\ell_p,~p\leq 1$ norm and the total variation (TV) regularizer \cite{Osher2005} to respectively enhance the magnitude of those dominated scattering points and the boundaries or edges in the image by exploiting their sparsity priors in the underlying image. Accordingly, their framework will suppress the sidelobes and produce a point enhanced or a piecewise smooth region enhanced image. It has been further evaluated that such enhancement can improve target recognition performance as expectation \cite{cetin2003feature,Kelly2012}. Additionally, they give an empirically conclusion that their regularization framework is also robust to the partially sampled data as well as observation noise \cite{cetin2003feature,franklin1974tikhonov}. This conclusion can be also confirmed and coincident with the latter emerging theory called compressive sampling or compressed sensing \cite{baraniuk2007compressive,donoho2006compressed,candes2006robust,Candes2011}, which theoretically demonstrates the overwhelming possibility of exactly recovering a sparse vector or a low rank matrix from its partially and randomly sampled entries\footnote{More precisely, recovering a low rank matrix with its partially known entries is referred to as matrix completion problem.}. Over past decade, this novel theory brings a new road to SAR or inverse SAR (ISAR) imaging by exploiting the priors of sparsity in an image for the sake of relieving the sampling burden and achieving an apparent improvement on image quality and resolution \cite{TelloAlonso2010,Patel2010,Samadi2011,Herman2008,Herman2009,Baraniuk2007,Zhang2010,Zhao2016}. More details about these sparsity-driven SAR imaging algorithms can refer to the surveys and references therein \cite{Potter2010,Cetin2014}.
\subsection{Motivation}
\par The above mentioned SAR imaging algorithms however only exploit the simple sparsity priors of several low level visual features in the image \cite{Smith1997}, which are insufficient and inaccurate to describe the complicated image contents and the target. Therefore, many researchers gradually concentrate on other priors and features. Wang \emph{et al.} \cite{wang2014enhanced} propose a target enhanced ISAR imaging algorithm based on Bayesian compressed sensing framework \cite{Ji2008} by exploiting the prior of continuity for the scatterers in the target scene. Additionally, Wang \emph{et al.} further leverage the Markov prior to encourage the continuity features \cite{Wang2015}. However, all above mentioned imaging algorithms are still unable to distinguish different semantic concepts in the image due to the gap between bottom level features and top level information in perception tasks \cite{Feng2016}. {Such gap is caused since the bottom level features such as points and edges are normally shared in different types of semantic objects. These shared low level features will firstly construct more complex structures as the mid-level features and finally form different semantic objects in the top level with different hierarchical approaches \cite{Bengio2009a}. As a consequence, due to the loss of the high level semantic information, these sparsity-driven imaging algorithms will enhance/suppress all strong/weak scattering points and edges of both the targets of interest and the unconcerned background clutters without considering their diverse semantic labels.} On one hand, from the ATR perspective, we generally pay more attention to those scattering points belonging to the interested target. In this regard, we only hope to enhance the points with the ¡°target¡± tag while suppressing others to obtain a tangible target enhanced image with a prominent target to clutter ratio. On the other hand, from the data acquisition perspective, the amount of target pixels is much smaller than that of the complete image. Only reconstructing these target pixels will be consequently able to reduce the required observations, which will further relieve the burden in data acquisition and sampling. These two perspectives straightforward raise a suggestion for the ATR task that whether we could directly reconstruct an only target image instead of throwing the unconcerned background clutters away after fully reconstruction as the conventional way does. To this end, we exploit another two target priors to estimate the position of the target region \cite{Wen2013} and reconstruct a specified number of scattering points within this region. But this algorithm heavily relies on a man-made target template in order to obtain an initial position estimation, which is not robust in practical environment and the issue of the semantic gap is not concerned as well.
\subsection{Main Contributions}
\par Motivated by the above analysis and explicitly addressing the semantic gap during SAR imaging procedure, we propose a novel semantic information guided iterative regularization framework for target oriented high resolution SAR image formation, which aims at enhancing the target while suppressing the background clutters. The main contributions of our framework are summarized as following.
\begin{itemize}
  \item We develop a semantics-specific regularizer for image formation by leveraging the statistical features of different semantic contents in a target scene image.
  \item  In order to determine the semantic label for each pixel in the image, we induce a prior-driven regularizer and some semantic causal rules from the high level semantic prior knowledge.
  \item Our regularized framework for image formation is further derived as a simple iteratively reweighted $\ell_1$ minimization problem, which makes the optimization easier to be implemented.
\end{itemize}
The extensive experimental results on the public MSTAR database\footnote{https://www.sdms.afrl.af.mil/index.php?collection=mstar} demonstrate the effectiveness and superiorities of our proposed framework in terms of target enhancement and clutters suppression, compared with the other state of the art imaging algorithms.
\par The remainder of this paper is organized as follows. Sec. \ref{Sec:Imagingmodel} firstly reviews the basic imaging model for the spotlight mode SAR, and then we propose our target oriented SAR image formation framework in Sec. \ref{Sec:Mainwork}. A detailed optimization scheme for the proposed framework is derived in Sec. \ref{Sec:Optimization} followed by an analysis of its convergence and computational complexity. Experiments are conducted in Sec. \ref{Sec:Experiments} and we conclude this paper in Sec. \ref{Sec:Conclusion}.

\section{Spotlight Mode SAR Imaging Model}\label{Sec:Imagingmodel}
\par
\begin{figure}
    \centering

\subfigure[]{\includegraphics[width=0.23\textwidth]{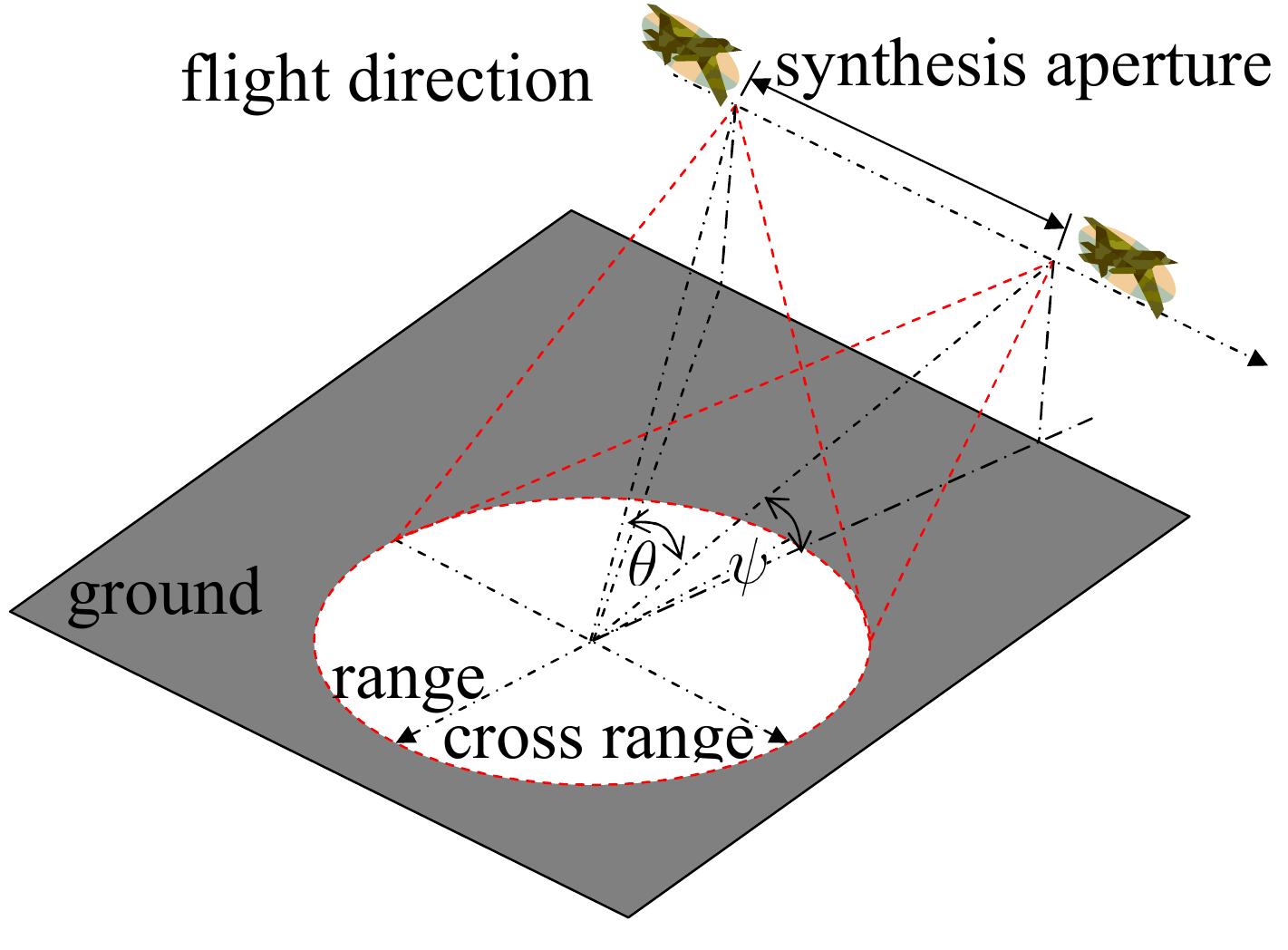}
 \label{fig_Spotlight_3D}}
\subfigure[]{\includegraphics[width=0.23\textwidth]{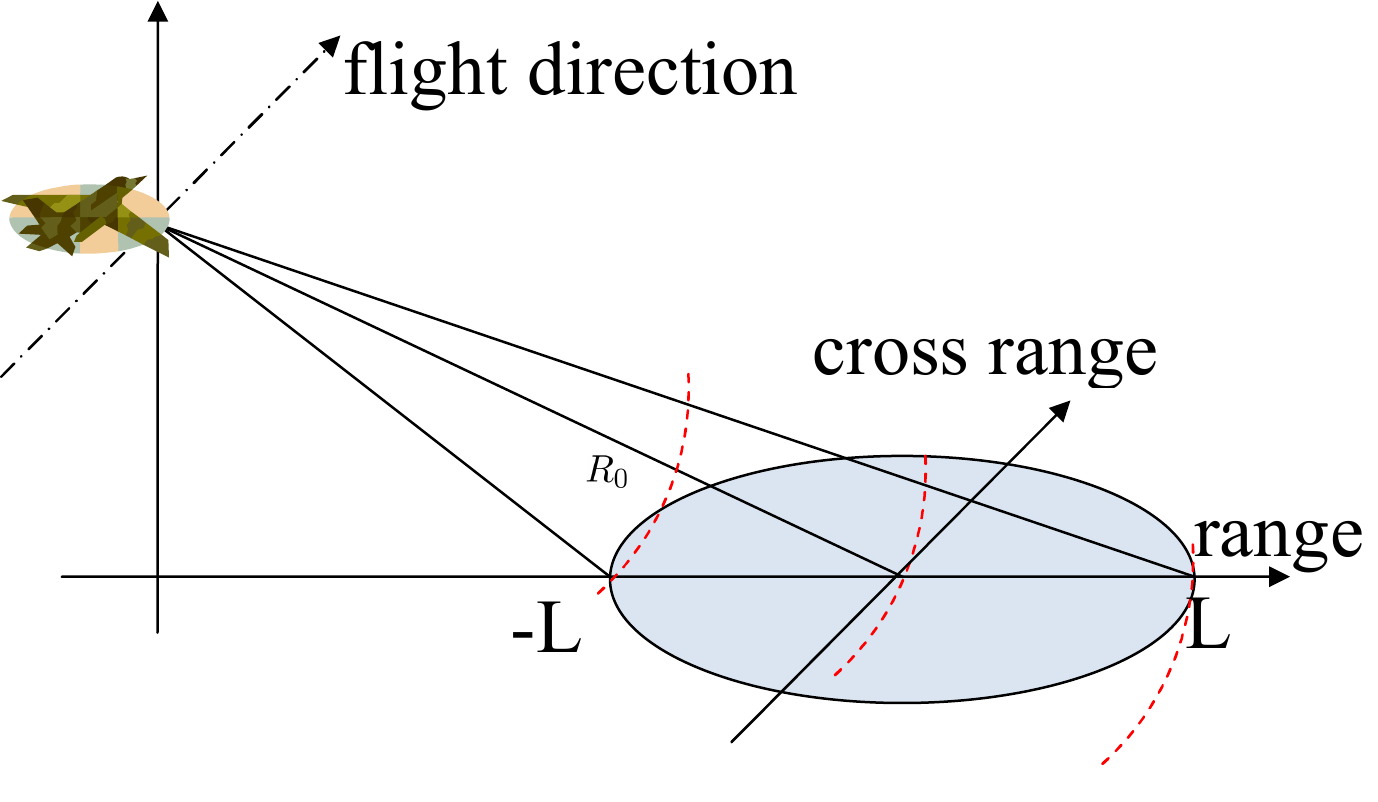}
 \label{fig_spotlight_side}}

\caption{The geometry of spotlight mode SAR.}
\label{Geo3D_SpotSAR}
\end{figure}

 In this section, we will review the basic observation model of the spotlight SAR to introduce the regularized imaging formulation \cite{Carrara1995}\cite{cetin2001feature}. Suppose the spotlight mode SAR stares a detected target scene by transmitting a series of linear frequency modulated pulse signal
\begin{equation}\label{Equ:LinearPulse}
 { s(t)=\exp\{j(\omega_0 t+\alpha t^2)\},~t\in[-\frac{T_p}{2},\frac{T_p}{2}],}
\end{equation}
where {$\omega_0$ denotes the carrier radian frequency}, $2\alpha$ is the so-called chirp rate, $t$ is the fast time and $T_p$ denotes the pulse width. Each transmitted pulse impinges the ground region and is subsequently reflected and received by the SAR platform. The geometry of this procedure is depicted in Figs. \ref{Geo3D_SpotSAR}, where $R_0$ denotes the distance between the SAR platform and the center of region, $\psi$ is the depression angle measuring the angle between the scene and the line of sight, $\theta$ is the total viewing angle extent and $L$ is the radius of the scene patch.
{After being demodulated, the observed echo signal from the $i$-th viewing angle $\theta_i$ can be formulated as \cite{Samadi2011}}
\begin{equation}\label{Equ:ContinuousFun}\small
\begin{aligned}
r(t,\theta_i)=\oint_{S(x,y)} g(x,y)\exp\{-j\Omega(t)(x\cos(\theta_i)+y\sin(\theta_i))\}dxdy,\\
t\in[-\frac{T_p}{2}+\frac{2(R_0+L)}{c},\frac{T_p}{2}+\frac{2(R_0-L)}{c}],\\
S(x,y)=\{(x,y)|x^2+y^2\leq L^2\},\\
\end{aligned}
\end{equation}
where $\Omega(t)=\frac{2}{c}(\omega_0+2\alpha(t-\frac{2R_0}{c}))$ represents the radial spatial frequency, $g(x,y)$ stands for the two dimensional microwave reflectivity density function of the illuminated target scene, namely the unknown SAR image and $c$ is the speed of light. It is worth noting that $r(t,\theta_i)$ is actually a finite slice of the two-dimensional (2D) Fourier transform of $g(x,y)$. During the flight, SAR platform keeps on staring the target region and receives a collection of $r(t,\theta_i)$ from $K$ different viewing angles $\{\theta_i\}_{i=1}^K$. Then we can reformulate Eq. \eqref{Equ:ContinuousFun} into a more compact form as
\begin{equation}\label{Equ:2Dfft}
  [{r}(t,\theta_1),\dots,{r}(t,\theta_K)]=\mathcal{A}(g(x,y))
\end{equation}
where $\mathcal{A}(\cdot)$ is a two dimensional Fourier transformation type operator. For each continuous signal $r(t,\theta_i)$, if it is sampled by a measurement operator $\Phi_i$ with the requirement of Nyquist sampling raw \cite{Shannon1949}, we will obtain an $N$-dimensional complex phase history vector  $\mathbf{r}_{\theta_i}=\Phi_i(r(t,\theta_i))\in\mathbb{C}^N$. For all viewing angles, a cascade discrete phase history vector $\mathbf{{r}}$ will be stacked as
\begin{equation}\label{Equ:Observationmodel}
  {\mathbf{r}}=\left[\begin{array}{c}
\mathbf{r}_{\theta_{1}}\\
\vdots\\
\mathbf{r}_{\theta_{K}}
\end{array}\right]=\left[\begin{array}{ccc}
\Phi_{1}\\
 & \ddots\\
 &  & \Phi_{K}
\end{array}\right]\left(\begin{array}{c}
r(t,\theta_{1})\\
\vdots\\
r(t,\theta_{K})
\end{array}\right).
\end{equation}
Then, combining Eq. \eqref{Equ:2Dfft} and \eqref{Equ:Observationmodel}, the discrete version of SAR observation model can be formulated as:
\begin{equation}\label{Equ:ImagingModel}
  {\mathbf{r}}=\mathbf{A}(\mathbf{G})+\mathbf{e},
\end{equation}
where $\mathbf{G}\in\mathbb{C}^{N\times K}$ is the discrete complex SAR image of the underlying target scene and $\mathbf{g}=\mathrm{vec}(\mathbf{G})$ is its vector form. $\mathbf{A}:\mathbb{C}^{N\times K}\rightarrow \mathbb{C}^{NK}$ is a linear discrete approximation of $\mathcal{A}$ as well as the sampling operator and $\mathbf{e}$ is the additive measurement noise. With the formulation \eqref{Equ:ImagingModel}, SAR image reconstruction can be regarded as estimating $\mathbf{G}$ from ${\mathbf{r}}$, which conventionally relies on the inverse or adjoint operator of $\mathbf{A}$ in the case of a small noise and the perfectly sampled data. In many practical situations, the received data will however be undersampled due to the limitation of bandwidth so that only partial entries of $\mathbf{r}$ are observed by $\mathbf{\widetilde{r}}=\Phi \mathbf{r}$, where $\Phi:\mathbb{C}^{NK}\rightarrow \mathbb{C}^M,~M\ll NK$ is the down sampling matrix. In this situation, those conventional inverse operator based imaging algorithms will not perform well as Eq. \eqref{Equ:ImagingModel} becomes ill-posed. Alternatively, the principle way to deal with this problem is to incorporate some constraint on the solution by regularization. Then a desired solution will be obtained from the following regularized linear inverse problem \cite{franklin1974tikhonov}.
\begin{equation}\label{Equ:ImageReconstruction}
  \min_{\mathbf{G}} \lambda\Psi(\mathbf{G})+\frac{1}{2}\|{\mathbf{\widetilde{r}}}-\Phi\mathbf{A}(\mathbf{G})\|_2^2.
\end{equation}
where $\Psi(\mathbf{G})$ is a regularization function imposing some required properties on $\mathbf{G}$, $\lambda$ is the regularization parameter and the second term is called the data fidelity. From the Bayesian inference aspect, the imaging problem can be also interpreted as an estimation of a latent variable $\mathbf{G}$ given the observed variable $\widetilde{\mathbf{r}}$ and  it can be formulated as the following maximum a posterior (MAP) estimator \cite{wang2014enhanced}\cite{Ji2008}.
\begin{equation}\label{Equ:MAPImaging}
\begin{split}
  \arg\max_{\mathbf{G}} p(\mathbf{G}|\mathbf{\widetilde{r}})\propto p(\mathbf{\widetilde{r}}|\mathbf{G})p(\mathbf{G})\\
  =\arg\min_{\mathbf{G}} -\log p(\mathbf{\widetilde{r}}|\mathbf{G})-\log p(\mathbf{G}),
\end{split}
\end{equation}
where $p(\widetilde{\mathbf{r}}|\mathbf{G})$ is the likelihood function of $\mathcal{L}(\mathbf{G}|\widetilde{\mathbf{r}})$ given the outcome $\mathbf{\widetilde{r}}$ and $p(\mathbf{G})$ is the priori distribution of latent $\mathbf{G}$. When the multivariate Gaussian measurement noise with zero mean vector is considered in the previous algorithms, $-\log p(\mathbf{\widetilde{r}}|\mathbf{G})$ will be equivalent to the data fidelity in \eqref{Equ:ImageReconstruction} and $\lambda^{-1}$ can control the noise precision. The remaining term $-\log p(\mathbf{G})$ will be described by the regularization function in \eqref{Equ:ImageReconstruction}, namely $\Psi(\mathbf{G})$. In the next section, this term will be specially designed to meet our requirement.
\section{Semantic Information Guided Target Oriented High Resolution SAR Image Formation}\label{Sec:Mainwork}
In this section, we will explicitly address the problem of semantic gap in the imaging process and develop a target oriented SAR image formation framework, which aims at enhancing the target while suppressing the background clutters in the generated image. To implement this task, two core issues should be taken into consideration, i.e., 1) how to determine the semantic labels for every {pixel} in an unknown SAR image without available supervised training data and 2) which regularizer can be exploited to simultaneously enhance and suppress pixels according to their semantic labels. In the following subsections, we will address these two issues explicitly.
\subsection{Semantics-Specific Regularizer with Statistical Features}\label{subsec:regularizer}
\par According to the above discussion, sparsity-driven imaging algorithms exploit the sparse priors of the pixel magnitude and edge or contour feature in an underlying SAR image. In this case, $\Psi(\mathbf{G})$ in \eqref{Equ:ImageReconstruction} is chosen as the $\ell_1$ norm and TV regularizer to regularize each pixel and edge independently without concerning its semantic label. In order to involve the semantic information, it is supposed that the magnitude of each pixel $g_s$ will be conditioned on its semantic label $y_s$ and several semantic features in $\Theta$. When the label set is finite and it contains $C$ categories, all pixels in $\mathbf{G}$ will be correspondingly clustered into $C$ disjoint groups based on their semantic labels. By assuming the conditional independence of each pixel on its label $y_s$ as well as $\Theta$ within each group, we have
\begin{equation}\label{Equ:Classcondition}
  p(\mathbf{G}|\mathbf{Y},\Theta)=\Pi_{c=1}^C\Pi_{s}^{N_c}p({g}_{s}|{y}_{s}=c,\Theta_c),
\end{equation}
where $y_s=c$ means that the semantic label of $g_s$ is $c$, $N_c$ computes the total number of pixels in class $c$ and $\Theta_c$ is the semantic feature set of this class. In this paper, three semantic labels in the target scene are concerned, including  ``background", ``target'' as well as ``shadow", namely $c\in\{``b",``t",``s"\}$. Nevertheless, ``shadow" actually is only an auxiliary label for target inference, which will be discussed in Sec. \ref{subsec_labelinference}.
\par Up to now, the important issue to design our semantics-specific regularizer becomes to determine $p({g}_{s}|{y}_{s}=c,\Theta_c)$. Intuitively, we could choose different distribution functions for the pixels with different labels, but for the sake of simplicity we assume all pixels are drawn from independent identical distributions (i.i.d) with distinct statistical features. Accordingly, various distributions can be investigated \cite{Gui2010}, such as generalized Gamma distribution \cite{Zhang2015}, Weibull distribution \cite{Weisenseel1999}, $G^0$ distribution \cite{feng2013multiphase} etc, in which the distribution parameters will be served as the statistical features. In our framework, we will exploit the Gamma distribution to model the pixel magnitude as follows.
\begin{equation}\label{Equ:Gamma}
\begin{split}
  p({g}_{s}|{y}_{s}=c,\Theta_c=\{a_c,b_c\})=\frac{1}{\Gamma(a_c)b_c^{a_c}}|g_s|^{a_c-1}e^{-\frac{|g_s|}{b_c}},\\
  a_c>0,~b_c>0,
  \end{split}
\end{equation}
where $\Gamma(\cdot)$ is the gamma function, $a_c$ and $b_c$ are shape and scale statistical features for class $c$, respectively and $|g_s|$ is the magnitude of pixel $g_s$. To validate the compatibility of Gamma distribution compared with Rayleigh and Weibull, we choose a sample target SAR image from MSTAR database and fit the histograms of different semantic regions shown in Fig. \ref{fig_illustratetargetshadow}, where pixels belonging to the complete target and shadow are manually bounded by the red and blue dash boxes, respectively and the rest pixels belong to the background. It can be empirically observed from Figs. \ref{fig_background}, \ref{fig_target} and \ref{fig_shadow} that the histograms for different semantic regions are indeed the most coincident with the Gamma distribution.

\begin{figure*}
  \centering
  \includegraphics[width=0.8\textwidth]{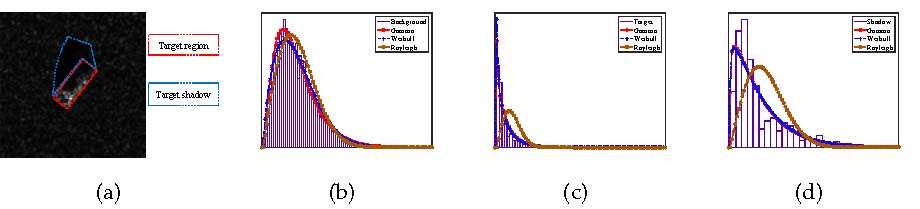}\\
  \caption{Illustration of the distributions fitting for different semantic groups.  (a) primary magnitude SAR image, (b) the background, (c) the complete target's region and (d) shadow.}\label{Fig_illustratetargetshadow}
\end{figure*}

\par Combining Eq. \eqref{Equ:Gamma} with \eqref{Equ:Classcondition} and taking its negative logarithmic form, we have:
\begin{equation}\label{Equ:neg_log}
\begin{split}
  -\log p(\mathbf{G}|\mathbf{Y},\Theta)
  &=\sum_{c:\forall y_s=c}\sum_s^{N_c}(\log(\Gamma(a_c))+a_c\log(b_c)\\
  &+\frac{1}{b_c}|g_s|+(1-a_c)\log(|g_s|+\epsilon))\\
  &=\sum_{c:\forall y_s=c}(N_c\log(\Gamma(a_c))+N_ca_c\log(b_c)\\
  &+\frac{1}{b_c}\sum_s^{N_c}|g_s|+(1-a_c)\sum_s^{N_c}\log(|g_s|+\epsilon)),
 \end{split}
\end{equation}
where $\epsilon>0$ is a small constant which is introduced to avoid infeasible in the case of $|g_s|=0$. With this formulation, if we leave out the irrelevant terms with respect to $g_s$, our semantics-specific regularizer will be designed as
\begin{equation}\label{Equ:LabelRegularizer}\small
  \Psi(\mathbf{G}|\mathbf{Y},\Theta)=\sum_{c:\forall y_s=c}\left(\frac{1}{b_c}\sum_s^{N_c}|g_s|+(1-a_c)\sum_s^{N_c}\log(|g_s|+\epsilon)\right).
\end{equation}
We can see that this regularizer is explicitly determined by the semantic labels and features $a_c,~b_c$. By choosing different features, the gamma distributions with distinct shapes and scales will be required to capture different semantic classes, which we are able to simultaneously enhance the target pixels and suppress others and this issue will be discussed in Sec. \ref{subsec:IRWL1}. {Compared with the traditional $\ell_1$ regularization, our $\Psi(\mathbf{G}|\mathbf{Y},\Theta)$ derived from the semantic related statistical distributions, comprises a weighted $\ell_1$ term $\frac{1}{b_c}\sum_s^{N_c}|g_s|$ and a sum of the logarithms term $(1-a_c)\sum_s^{N_c}\log(|g_s|+\epsilon)$, in which the semantic labels $y_s$ and features $a_c$, $b_c$ are specifically involved to control the distributions of different semantic classes. On the contrary, traditional $\ell_1$ norm or TV regularizer only derived from the simple bottom level priors treats every class equally without any distinction. To our best knowledge, this semantics-specific regularizer is initially developed in image reconstruction field.}  In particular, when $\forall~a_c=1$ and $b_c$ are identical for all classes, $\Psi(\mathbf{G}|\mathbf{Y},\Theta)$ becomes the normal $\ell_1$ regularization.
\subsection{Prior-Driven Regularizer with High level Semantic Information}\label{subsec_labelinference}
\par Assigning each pixel in the image with a proper candidate semantic label is generally referred to as the image semantic segmentation or pixel annotation which is another fundamental perception {task} in SAR image processing \cite{Krylov2011,Ghinelli1997The}. However, most of the algorithms addressing this problem are based on some classification strategies, where a large number of labeled pixels are required to train a classifier in a supervised way. Otherwise, only pixels clustering or image segmentation can be implemented \cite{Zhang2008}. In our imaging scenario, the pixels in the underlying SAR image are unknown so that no pixel can be labeled as the training data. {To address this problem, we have to incorporate some high level semantic priors and side information to infer the semantic labels $\mathbf{Y}$ in an unsupervised way and these semantic prior knowledge and side information can be readily obtained and induced from human's perceptive and cognitive experiences or some common senses.}
\begin{figure}
  \centering
  \includegraphics[width=0.49\textwidth]{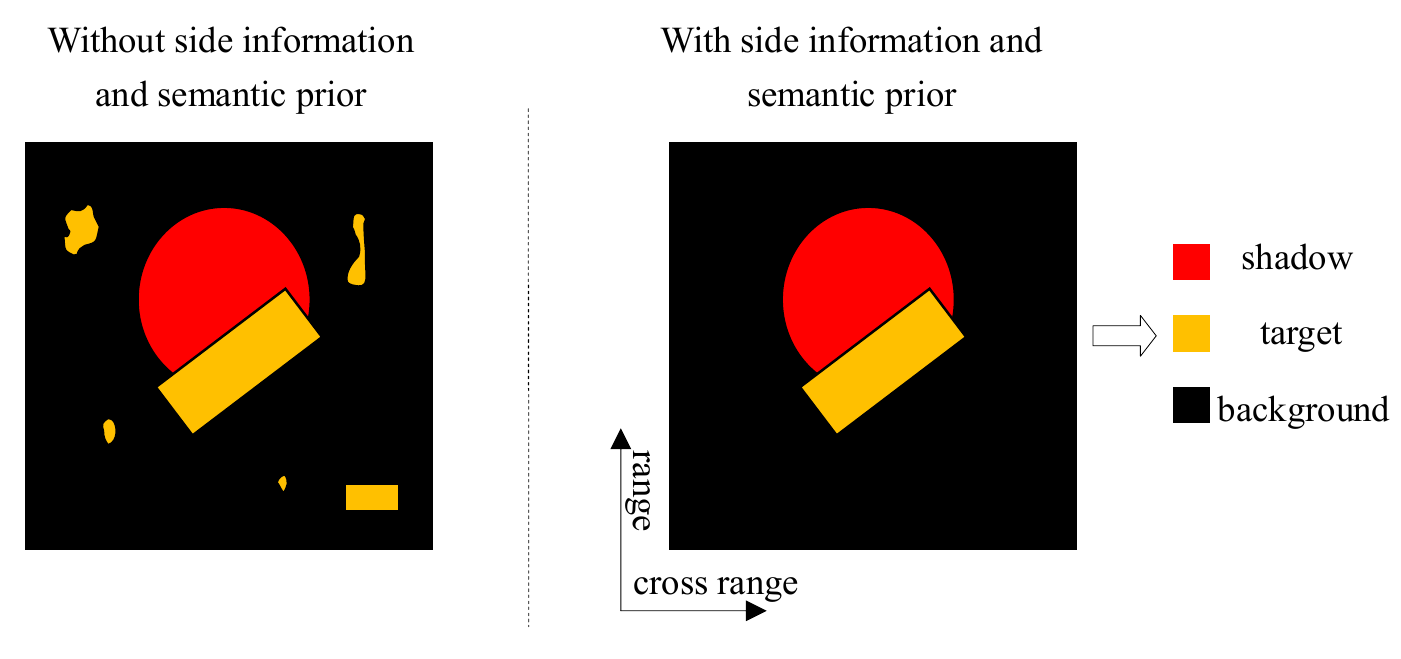}\\
  \caption{Illustration of side information and class-specific prior guided semantic segmentation. Left: the unsupervised segmentation result without any extra information. Right: the modified result involving the side information and class-specific semantic prior.}\label{Fig:SemanticIll}
\end{figure}
\par To establish the role of high level semantic priors and side information for label inference, we will firstly consider the following example. Suppose we have obtained a segmentation result of a SAR image with some unsupervised clustering algorithms shown in Fig. \ref{Fig:SemanticIll} (left). For a computer without providing it any other information of the target, shadow as well as background, it is not possible to assign the correct semantic label to each coloured segment. On the contrary, an expert who has known the side information of range and cross range direction as well as radar flight state is able to make an inference as following. Firstly, according to the principle of the spotlight mode SAR imaging, the shadow will locate in the downrange of the target when the radar is operated in side-looking mode \cite{Carrara1995}. According to the semantic prior knowledge, ``target" and ``shadow" always belong to a co-occurrence semantic pair in range direction so that the expert can state that target will locate beneath its shadow if range direction is established in Fig. \ref{Fig:SemanticIll} (right). Next, the man-made target will be continuous \cite{wang2014enhanced} and has an approximately regular geometry shape while other unconcerned natural objects or clutters are irregular in general. {For example, a man-made tank or a building can be modeled with a regular rectangle in a high resolution SAR image \cite{Ainsworth2008,Hummel2000}.} Through this shape prior, many unconcerned objects with irregular shapes can be excluded from the target of interest. Finally, side information about the volume of the target and the image resolution can be known in advance which can be served as the size semantic prior during target inference. Depending on this prior, the size of a target region in the SAR image can be approximately computed so that those oversized or undersized segments can be excluded. Accordingly, we will induce several causal rules with respect to the target from these high level semantic priors, namely 1) target pixel should appear beneath its shadow one, {2) target region should be continuous with an appropriate size and regular geometry shape}\footnote{Actually, this paper does not consider the geometry shape for simplicity but this prior has already widely exploited in other applications.}. These two rules can straightforwardly bridge the gap between each semantic concept and the segment, yielding the modified segmentation result associated with the estimated labels in Fig. \ref{Fig:SemanticIll} (right).
\par This example explicitly illustrates the superior effect of these high level semantic priors in label inference, where ``shadow" served as an auxiliary semantic tag will help us to locate the target. Following the above inference process, we will infer the semantic label by means of the random field model. Let $\mathcal{X}$ be the hidden image magnitude random field which is defined on a graph with the first order four neighbouring system $\mathcal{N}$ and any a recovered magnitude SAR image $|\mathbf{G}|\in\mathcal{X}$ is viewed as a configuration. Correspondingly, another hidden semantic label field $\mathcal{Y}$ is constructed for $\mathcal{X}$ with a label configuration $\mathbf{Y}$. Firstly, considering the continuity prior, it suggests that pixels with the same semantic labels will generally cover a continuous region with a high probability \cite{Wang2015,Wen2013,wang2014enhanced}. This prior property can be described by modeling $\mathcal{Y}$ as a Markov random field (MRF) \cite{Li2009}. According to the equivalence between Gibbs and MRF established by Hammersley-Clifford theorem, $p(\mathbf{Y})$ can be formulated with the following form
\begin{equation}\label{Equ:MRF}
  p(\mathbf{Y}|\beta)=\frac{1}{Z_y(\beta)}e^{-\beta E(\mathbf{Y})}=\frac{1}{Z_y(\beta)}e^{-\beta\sum_{c\in\mathcal{C}}\varphi_c(\mathbf{Y})},
\end{equation}
where $Z_y(\beta)=\sum_{\mathbf{Y}} e^{-\beta E(\mathbf{Y})}$ is the partition function for normalization with a parameter $\beta$ and $E(\mathbf{Y})=\sum_{c\in\mathcal{C}}\varphi_c(\mathbf{Y})$ is referred to as the energy function computing a sum of clique potential $\varphi_c(\mathbf{Y})$ over all cliques set $\mathcal{C}$. In this paper, we only consider the pairwise potential function $\varphi(y_s, y_t),~t\in\mathcal{N}_s$ to encourage the label consistency between the node \emph{s} and its neighbours $t\in\mathcal{N}_s$. In the previous researches, many models have been investigated to construct this function such as homogenous Potts or more generally Ising model equipped with a simpler Hamiltonian function \cite{Boykov2001Fast}. Nevertheless, these functions do not take the other semantic priors into account. Therefore, we specially design a semantic related pairwise function $\varphi(y_s, y_t)$ by further incorporating the co-occurrence prior. More concretely, in addition to encouraging the adjacent pairwise labels to be consistent, our function will impose different penalties according to the previous induced semantic causal rules, i.e., a larger penalty on those pairs violating the co-occurrence rule, and vice versa. Our final designed pairwise potential function is established in Table \ref{Tab:PairwiseSemantic}, where the notation $``s|t" $ and $``\frac{t}{s}"$ stand for the spatial position between node ``s" and ``t" \cite{Weisenseel1999} in the image. We can observe from this function that a larger penalty is imposed on those pairs that ``target" is above the ``shadow" and ``background" is above the ``target". Then the prior-driven regularizer for semantic labels can be derived from $-\log p(\mathbf{Y}|\beta)$ by leaving out the irrelevant constant term with respect to $\mathbf{Y}$. The remaining causal rules will be considered during label inference process, which will be discussed in Sec. \ref{Sec:Optimization}.
\begin{table}
\caption{The pairwise potential function $\varphi(y_{s},y_{t})$}\label{Tab:PairwiseSemantic}
\centering
\subtable[$s(t)|t(s)$]{
       \begin{tabular}{|c|c|c|c|}
\hline
 & $s$ & $b$ & $t$\tabularnewline
\hline
$s$ &0  & 1 & 1\tabularnewline
\hline
$b$ & 1 & 0 &1\tabularnewline
\hline
$t$ &1  & 1 & 0\tabularnewline
\hline
\end{tabular}
       \label{tab:firsttable}
}
\subtable[$\frac{s}{t}$ ]{
       \begin{tabular}{|c|c|c|c|}
\hline
 & $s$ & $b$ & $t$\tabularnewline
\hline
$s$ &0& 2 &1 \tabularnewline
\hline
$b$ & 1 & 0 & 2\tabularnewline
\hline
$t$ & 2 & 1 &0\tabularnewline
\hline
\end{tabular}
}
\subtable[$\frac{t}{s}$ ]{
       \begin{tabular}{|c|c|c|c|}
\hline
 & $s$ & $b$ & $t$\tabularnewline
\hline
$s$ &0& 1 &2 \tabularnewline
\hline
$b$ & 2 & 0 & 1\tabularnewline
\hline
$t$ & 1 & 2&0\tabularnewline
\hline
\end{tabular}
}
\end{table}
\subsection{Target-Enhanced Image Formation via Iteratively Reweighted $\ell_1$ Minimization}\label{subsec:IRWL1}
\par In the previous subsections, we have developed the  regularizer $\Psi(\mathbf{G}|\mathbf{Y},\Theta)$ by exploiting the statistical distribution features for different semantic class and the label prior function $p(\mathbf{Y}|\beta)$ by incorporating several high level semantic priors. The rest issue concentrates on developing a computational framework of enhancing the target while suppressing others.
\par Let us review our imaging framework from Bayesian perspective, which we aim at simultaneous estimating a SAR image $\mathbf{G}$, its latent label configuration $\mathbf{Y}$ and a set of class-specific features $\Theta$ from the partially observed phase history $\widetilde{\mathbf{r}}$. Therefore, deriving from Eq. \eqref{Equ:MAPImaging} and considering the Eqs. \eqref{Equ:MRF}, \eqref{Equ:neg_log}, the final target oriented SAR image formation will be accordingly formulated as the following MAP
\begin{equation}\label{Equ:FinalReconstruction}
\begin{split}
  &\arg\max p(\mathbf{G},\mathbf{Y},\Theta|\widetilde{\mathbf{r}})\\
=&\arg\min -\log p(\widetilde{\mathbf{r}}|\mathbf{G})-\log p(\mathbf{G}|\mathbf{Y},\Theta)-\log p(\mathbf{Y}|\beta)\\
=&\arg\min \frac{1}{2\lambda}\|\widetilde{\mathbf{r}}-\Phi\mathbf{ A}(\mathbf{G})\|_2^2+\Psi(\mathbf{G}|\mathbf{Y},\Theta)\\
  +&\sum_{c:\forall y_s=c}\left( N_c\log(\Gamma(a_c))+N_ca_c\log(b_c)\right)+\beta\sum_s\sum_{t\in\mathcal{N}_s}\varphi(y_s,y_t).
  \end{split}
\end{equation}
This optimization problem can be alternatively solved by optimizing $\mathbf{G}$, $\mathbf{Y}$ and $\Theta$ with the following iterative scheme:
\begin{equation}\label{Equ:Optscheme}
\begin{split}
  \mathbf{G}^{(k+1)}&\gets \arg\min_\mathbf{G} \frac{1}{2\lambda}\|\widetilde{\mathbf{r}}-\Phi\mathbf{ A}(\mathbf{G})\|_2^2+\Psi(\mathbf{G}|\mathbf{Y}^{(k)},\Theta^{(k)}),\\
  \mathbf{Y}^{(k+1)}&\gets \arg\min_\mathbf{Y} \sum_{c:\forall y_s=c}\left( N_c\log(\Gamma(a_c^{(k)}))+N_ca_c\log(b_c^{(k)})\right)\\
  +&\Psi(\mathbf{G}^{(k+1)}|\mathbf{Y},\Theta^{(k)})+\beta\sum_s\sum_{t\in\mathcal{N}_s}\varphi(y_s,y_t),\\
  \Theta^{(k+1)}&\gets \arg\min_{\Theta} \sum_{c:\forall y_s^{(k+1)}=c}\left( N_c\log(\Gamma(a_c))+N_ca_c\log(b_c)\right)\\
  +&\Psi(\mathbf{G}^{(k+1)}|\mathbf{Y}^{(k+1)},\Theta),
  \end{split}
\end{equation}
which respectively corresponds to SAR image formation, semantic label inference and feature parameters update. Compared with previous sparsity-regularized imaging frameworks, our regularizer seems to be more sophisticated in form than traditional $\ell_1$ norm or TV regularizer. However, we next attempt to simplify the optimization by some choices of feature parameters.
\par Considering our semantics-specific regularizer for SAR magnitude image
\begin{equation}\label{Equ:Regularizer}\small
    \Psi(\mathbf{G}|\mathbf{Y},\Theta)=\sum_{c:\forall y_s=c}\left(\frac{1}{b_c}\sum_s^{N_c}|g_s|+(1-a_c)\sum_s^{N_c}\log(|g_s|+\epsilon)\right),
\end{equation}
when the shape feature $a_c=1,~\forall c$, this regularizer will become the weighted $\ell_1$ regularizer, in which the weight $\frac{1}{b_c}$ will be assigned to those pixels of class $c$ to control the magnitude scale of pixels within this group. When $a_c<1,~\forall c$, this regularizer will be a difference of convex (DC) function \cite{Tao1998} or the convex-concave function \cite{Schuele2005}. A common strategy to deal with this problem is utilizing the majorization-minimization (MM) algorithm. Let $f(\mathbf{x})=(1-a_c)\sum_i\log(|x_i|+\epsilon)$ be the concave function in \eqref{Equ:Regularizer}. MM is an iterative algorithm which employs a series of surrogate functions $\widehat{f}(\mathbf{x}|\mathbf{x}^{(t)})$ at each iterative point $\mathbf{x}^{(t)}$ such that $\widehat{f}(\mathbf{x}|\mathbf{x}^{(t)})\geq f(\mathbf{x})$ and the equality holds when $\mathbf{x}=\mathbf{x}^{(t)}$. Then the next point will be obtained as $\mathbf{x}^{(t+1)}\gets \arg\min_{\mathbf{x}} \widehat{f}(\mathbf{x}|\mathbf{x}^{(t)})$, yielding the following decrease in $f(\mathbf{x})$
\begin{equation}\label{Equ:ConverMM}
  f(\mathbf{x}^{(t)})=\widehat{f}(\mathbf{x}^{(t)}|\mathbf{x}^{(t)})\geq\widehat{f}(\mathbf{x}^{(t+1)}|\mathbf{x}^{(t)})\geq f(\mathbf{x}^{(t+1)}).
\end{equation}
With the iteration proceeding, the function will be gradually minimized. In our case, the surrogate function of concave $f(\mathbf{x})$ at $\mathbf{x}^{(t)}$ can be set with its tangent function at current point as
\begin{equation}\label{Equ:Surrogate}
\begin{split}
  \widehat{f}(\mathbf{x}|\mathbf{x}^{(t)})=f(\mathbf{x}^{(t)})+<\nabla f(\mathbf{x}^{(t)}),\mathbf{x}-\mathbf{x}^{(t)}>\\
  =f(\mathbf{x}^{(t)})+(1-a_c)\sum_i\frac{|x_i|-|x_i^{(t)}|}{|x_i^{(t)}|+\epsilon},
  \end{split}
\end{equation}
where $<\cdot,\cdot>$ stands for the inner product operator. With this type of surrogate function, the sub-problem in \eqref{Equ:Optscheme} of image recovery  can be reformulated as  a series of iterative optimizations.
\begin{equation}\label{Equ:IterativeSAR}\small
\begin{split}
 \mathbf{ G}^{(t+1)}&\gets\arg\min_{\mathbf{G}} \frac{1}{2\lambda}\|\widetilde{\mathbf{r}}-\Phi\mathbf{ A}(\mathbf{G})\|_2^2\\
 &+\sum_{c:\forall y_s=c}\left(\frac{1}{b_c}\sum_s^{N_c}|g_s|+f({\mathbf{g}}_s^{(t)})+(1-a_c)\sum_s^{N_c}\frac{|g_s|-|g_s^{(t)}|}{|g_s^{(t)}|+\epsilon}\right)\\
 &=\arg\min_{\mathbf{G}} \frac{1}{2\lambda}\|\widetilde{\mathbf{r}}-\Phi\mathbf{ A}(\mathbf{G})\|_2^2\\
 &+\sum_{c:\forall y_s=c}\left((\frac{1}{b_c}+\frac{1-a_c}{|g_s^{(t)}|+\epsilon})\sum_s^{N_c}|g_s|\right)\\
 &=\arg\min_{\mathbf{G}} \frac{1}{2\lambda}\|\widetilde{\mathbf{r}}-\Phi\mathbf{ A}(\mathbf{G})\|_2^2+\sum_i w_i^{(t+1)}|g_i|,
 \end{split}
\end{equation}
where $w_i^{(t+1)}$ is the current weight for $i$-th entry in $\mathbf{G}$ with value
\begin{equation}\label{Equ:Weightssetting}
\begin{split}
  w_i^{(t+1)}=\frac{1}{b_c}+\frac{1-a_c}{|g_i^{(t)}|+\epsilon},\\
  \mathrm{s.t.}~y_i=c,~0<a_c< 1,\forall c\in\{``b",``t",``s"\}.
  \end{split}
\end{equation}
With above derivations, the first optimization problem in \eqref{Equ:Optscheme}, namely image recovery can be reformulated as a simple iteratively reweighted $\ell_1$ minimization (IRW-$\ell_1$) problem in the case of $a_c<1$ which can be readily solved by various existing solvers such as FISTA \cite{beck2009fast} and NESTA \cite{Becker2011}. We can also observe from \eqref{Equ:Weightssetting}, this weight setting also includes the case $a_c=1$, $w_i^{(t+1)}=\frac{1}{b_c}$. Accordingly, if the weights of the target pixels are always smaller than those from other classes as $w_i\leq w_j\leq w_k,~\forall y_i=``t", y_j=``b", y_k=``s"$, pixels excluding the target will be imposed on a larger penalty so as to generate a target enhanced image.
\subsection{Framework Interpretation and Analysis}\label{Subsec:Analysis}
\begin{figure}
  \centering
  \includegraphics[width=0.49\textwidth]{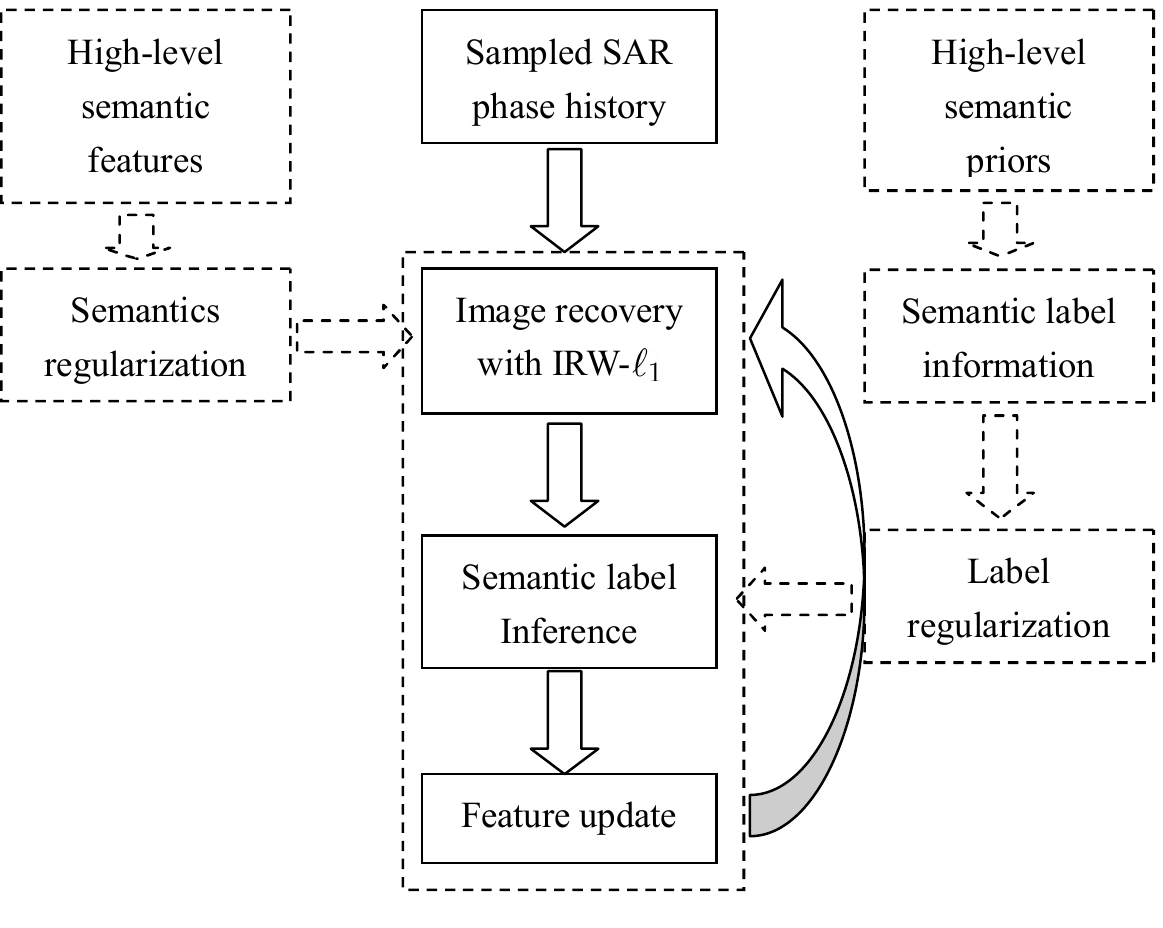}\\
  \caption{The framework of semantic information guided target oriented SAR image formation.}\label{Fig.Framework}
\end{figure}
\par According to above presentations, we have implemented the novel target oriented SAR image formation framework and we will illustrate this framework in Fig. \ref{Fig.Framework}. In this part, we will present an interpretation and analysis of the proposed framework to further highlight its superiorities.
\par Compared with the sparsity-driven imaging algorithms, we develop two new regularization functions, namely target-inducing $\Psi(\mathbf{G}|\mathbf{Y},\Theta)$ and semantic priors-inducing $\sum_s\sum_{t\in\mathcal{N}_s}\varphi(y_s,y_t)$ instead of the conventional sparsity-inducing $\ell_1$ norm and piecewise smoothing-inducing TV regularizer. Our first function is used to regularize the pixels to be recovered to have different desired statistical features according to their semantic labels. The second one imposes the regularization on the latent semantic labels, which not only provides a local continuity property but also obeys the high level co-occurrence rule among these labels. These two semantic information guided regularization functions enable us to recover a target enhanced SAR image as well as the label configuration without exploiting any training data. Considering the complexity of the proposed functions during optimization, we further reformulate the subproblem of SAR image formation as a simpler IRW-$\ell_1$ problem under some constraints, which is much easier to be addressed with a broad of existing solvers.
\par According to the discussion in the previous section, the performance of regularizer $\Psi(\mathbf{G}|\mathbf{Y},\Theta)$ is controlled by the latent semantic group structure as well as the desired features. A straightforward question naturally raises how to ensure a reliable label inference from the undersampled data especially when the undersampling rate is low. In our work, we alteratively reconstruct the image and update the labels, through which some false labels are expected to be adjusted in the subsequent iterations. Nevertheless, once the target pixels were labeled as the background or shadow, they would be always suppressed in the following iterations without refinement. To relieve such risk, we prefer a progressive suppression scheme by introducing a gradually increasing parameter $\lambda^{(k)}=\min\{e^{k}/\lambda_0,\lambda\}$ in \eqref{Equ:Optscheme}, where $k$ is the iteration, $\lambda_0$ is a constant controlling the increasing rate and $\lambda$ is the original required regularization parameter. More generally speaking, we impose a conditional prior distribution on $\lambda$ to control its increasing rate. With this setting, a slight suppression will be realized on pixels in the first few iterations so that some falsely estimated labels are capable of being gradually refined in the subsequent iterations \footnote{The requirement of exactly accurate estimation is always unrealistic so that the proposed strategy is only a feasible trick for relieve this problem. }. This strategy will be empirically validated in Sec. \ref{Subsec:paraanalysis} in detail and it works well in our experiments.  After the first few iterations, larger weights will be imposed on those background and shadow pixels to yield a stronger suppression and then their magnitudes will mostly tend to zeros. In this situation, the pixels with ``background" or ``shadow" labels will be gradually clustered into one group containing zero values, yielding the following two unexpected consequences. In the first case, these zero pixels will be assigned with ``background" label so that the ``shadow" set will be gradually empty, leading to a broken down in FCM. In the other case, they will be regarded as the ``shadow" pixels so that some target pixels with relatively small magnitude will be reassigned as the ``background" leading to an incomplete target region. To get rid of these problems, a possible trick is to cluster the pixels into two groups instead of three after several iterations, namely target and others.
\section{Optimization and Analysis}\label{Sec:Optimization}
The proposed iterative framework consists of three phases, namely image formation with IRW-$\ell_1$, semantic label inference and feature parameters update. In this section, we will present the detailed optimization schemes for each phase, respectively.
\subsection{Iteratively Reweighted $\ell_1$ Minimization for Image Recovery}
\par We have derived a IRW-$\ell_1$ framework for SAR image recovery, in which we solve a series of weighted $\ell_1$ minimization problem to estimate a current solution $\mathbf{G}$ and then update the weight matrix $\mathbf{W}$ iteratively. Therefore, the basic optimization in this subproblem will be a weighted $\ell_1$ minimization given as follows.
\begin{equation}\label{Opt:ImageRecov}
\begin{split}
 \min_{\mathbf{G}} \frac{1}{2\lambda}\|\widetilde{\mathbf{r}}-\Phi\mathbf{ A}(\mathbf{G})\|_2^2+\sum_i w_i|g_i|.
 \end{split}
\end{equation}
This problem can be solved by many existing solvers and it will be addressed with FISTA in this paper \cite{beck2009fast}. FISTA is a fast version of traditional iterative shrinkage thresholding algorithm (ISTA) by involving a Nesterov acceleration strategy \cite{Nesterov1983}, which can provide a significantly better global convergence rate and preserve the computational simplicity. This algorithm is proceeding with the following iterative scheme:
\begin{equation}\label{Sol:FISTA}
\begin{split}
  \widehat{\mathbf{G}}^{(t)}&=\mathcal{S}_{\lambda,\mathbf{W}}(\mathbf{G}^{(t)}-\frac{1}{L_{\widetilde{\Phi}}}\widetilde{\Phi}^*\widetilde{\Phi}(\mathbf{G}^{(t)})+\frac{1}{L_{\widetilde{\Phi}}}\widetilde{\Phi}^* (\widetilde{\mathbf{r}})),\\
  \alpha^{(t+1)}&=\frac{1+\sqrt{1+4(\alpha^{(t)})^2}}{2},\\
  \mathbf{G}^{(t+1)}&=\mathbf{\widehat{G}}^{(t)}+\frac{\alpha^{(t)}-1}{\alpha^{(t+1)}}(\mathbf{\widehat{G}}^{(t)}-\mathbf{\widehat{G}}^{(t-1)}),
  \end{split}
\end{equation}
where $\widetilde{\Phi}=\Phi \mathbf{A}$, $\widetilde{\Phi}^*$ is its adjoint operator, $\mathcal{S}_{\lambda,\mathbf{W}}(\mathbf{X})$ is element-wise shrinkage operator which is given by \eqref{Equ:Shinkage}, $L_{\widetilde{\Phi}}=\|\widetilde{\Phi}^{*}\widetilde{\Phi}\|$ and $\alpha^{(1)}=1$.
\begin{equation}\label{Equ:Shinkage}
  \mathcal{S}_{\lambda,\mathbf{W}}(\mathbf{X})=\frac{\max\{|\mathbf{X}|-\lambda\mathbf{W},\mathbf{0}\}}{\max\{|\mathbf{X}|-\lambda\mathbf{W},\mathbf{0}\}+\lambda\mathbf{W}}\odot\mathbf{X},
\end{equation}
where $\odot$ is the element-wise product (Hadamard product). FISTA converges when it reaches a stationary point and we can then update the corresponding weight matrix with Eq. \eqref{Equ:Weightssetting}.
\subsection{Semantic Label Inference}
\par The subproblem for semantic label inference can be summarized as
\begin{equation}\label{Opt:SemanticInference}
\begin{split}
\min_\mathbf{Y} \sum_{c:\forall y_i=c}\left( N_c\log(\Gamma(a_c))+N_ca_c\log(b_c)\right)+\Psi(\mathbf{G}|\mathbf{Y},\Theta)\\
+\beta\sum_i\sum_{j\in\mathcal{N}_i}\varphi(y_i,y_j).  \end{split}
\end{equation}
Considering the effectiveness and efficiency, we will adopt a fast greedy approach, termed as Iterated Conditional Modes (ICM) \cite{Besag1986}, whose central idea is to iteratively optimize a current active label with others fixed to obtain a local optimal solution. Due to the non-convexity of Opt. \eqref{Opt:SemanticInference}, the initial labels and the update order will generally have an impact on the solution. For semantic label initialization, we adopt the typical fuzzy c-means (FCM)  clustering algorithm on the current magnitude image $\mathbf{G}$ and obtain an initial unsupervised segmentation \cite{Bezdek1984}. To assign semantic labels to the obtained clusters, we conclude from the semantic prior that the man-made target usually contains the strong scattering points due to its regular geometry structure such as dihedral. Therefore, among all semantic labels, the mean magnitude and variance of the target clusters will be the largest while those of shadow pixels will be the smallest. According to this prior knowledge, the semantic labels can be initially assigned to each segment. During the optimization, we will choose a stochastic order for convenience and the optimization problem with respect to the $i$-th label will be
\begin{equation}\label{Equ:Single}
\begin{split}
y_i^{(t+1)}\gets \arg\min_{{y}_i}\beta\sum_{j\in\mathcal{N}(i)}\varphi(y_i, y_j^{(t)})-a(y_i)\log(\frac{|g_i|}{b(y_i)})\\
+\frac{|g_i|}{b(y_i)}+\log(\Gamma(a(y_i))),
  \end{split}
\end{equation}
where $a(y_i)=a_c$ and $b(y_i)=b_c$ if $y_i=c$. The above \eqref{Equ:Single} is a simple univariate and unidimensional optimization whose solution can be computed by checking the each value of $y_i$ that minimizes \eqref{Equ:Single}. It is worth pointing out that it will be much faster and efficient to synchronously optimize those nodes that are not adjacent. The algorithm stops when the number of variational labels between two iterations is lower than a predefined amount or the iteration reaches its maximum. After the algorithm stops, we empirically observe from the result that there always exists some isolated ``target" points or regions. To refine the result, the prior of the target size can be used to further make a decision, where only the region with an appropriate size are preserved while others are simply set as the ``background" label.
\subsection{Semantic Feature Parameters Update}\label{subsec:featureupdate}
The optimization problem with respect to two semantic features is formulated as:
\begin{equation}\label{Opt:ParaGamma}
\begin{split}
  \min_{a_c,b_c} \sum_{c,\forall y_s=c} (N_c\log(\Gamma(a_c))+N_ca_c\log(b_c)+\frac{1}{b_c}\sum_s^{N_c}|g_s|\\
  +(1-a_c)\sum_s^{N_c}\log(|g_s|+\epsilon)),~0<a_c\leq 1.
\end{split}
\end{equation}
After analyzing this optimization carefully, we find that $b_c$ admits a closed form solution given by
\begin{equation}\label{Opt:para_bc}
  b_c^*=\arg\min_{b_c} N_ca_c\log(b_c)+\frac{1}{b_c}\sum_s^{N_c}|g_s|=\mu_c/a_c,
\end{equation}
where $\mu_c$ is the mean magnitude of pixels from class $c$. Substituting \eqref{Opt:para_bc} into \eqref{Opt:ParaGamma}, we can obtain a simpler problem only with respect to $a_c$ as:
\begin{equation}\label{Opt:Paraac}
\begin{split}
  \min_{a_c} \sum_{c,\forall y_s=c} (N_c\log(\Gamma(a_c))+N_ca_c\log(\mu_c/a_c)+a_cN_c\\
  -a_c\sum_s^{N_c}\log(|g_s|+\epsilon)),~0<a_c\leq 1.
\end{split}
\end{equation}
According to the previous presentation in Sec. \ref{Sec:Mainwork}, these features directly determine the weight for pixels so that we have developed two types of constraints for the desired weights to yield a target enhanced SAR image. Considering the first type, we require that all weights of background pixels are no more than those of shadow while no less than those of target. It holds if and only if $\widehat{w_t}\leq \widetilde{w_b}$ and $\widehat{w_b}\leq \widetilde{w_s}$, where $\widehat{w_c}=\max_{i,y_i=c}|w_i|+\epsilon$ and $\widetilde{w_c}=\min_{i,y_i=c}|w_i|+\epsilon$. In this case, $\widehat{w_t}=\frac{a_t}{\mu_t}+\frac{1-a_t}{\widetilde{g_t}}$, $\widetilde{w_b}=\frac{a_b}{\mu_b}+\frac{1-a_b}{\widehat {g_b}}$, $\widehat{w_b}=\frac{a_b}{\mu_b}+\frac{1-a_b}{\widetilde{g_b}}$ and $\widetilde{w_s}=\frac{a_s}{\mu_s}+\frac{1-a_s}{\widehat {g_s}}$ result in the following constraints on $a_t$, $a_b$ and $a_s$, respectively.
\begin{equation}\label{Equ:Constraints}
  \begin{split}
  ~\max \{\frac{\widehat{w_t}\mu_b\widehat{g_b}-\mu_b}{\widehat{g_b}-\mu_b},\frac{\mu_b\widetilde{g_b}\widetilde{w_s}-\mu_b}{\widetilde{g_b}-\mu_b}\}\leq a_b\leq 1,\\
  \frac{\mu_t(\widetilde{g_t}\widetilde{w_b}-1)}{\widetilde{g_t}-\mu_t}\leq a_t \leq 1,~\frac{\mu_s(\widehat{g_s}\widehat{w_b}-1)}{\widehat{g_s}-\mu_s}\leq a_s\leq 1.
  \end{split}
\end{equation}
All above constraint sets for $a_t$, $a_b$ and $a_s$ should not be empty, we moreover have
\begin{equation}\label{Equ:Constraints_feasible}
  \begin{split}
  \max \{\frac{\mu_b(\widehat{g_b}-\mu_t)}{\mu_t(\widehat{g_b}-\mu_b)},\frac{\mu_b(\widetilde{g_b}-\mu_s)}{\mu_s(\widetilde{g_b}-\mu_b)}\}\leq a_b,\\
  \frac{\mu_t(\widetilde{g_t}-\mu_b)}{\mu_b(\widetilde{g_t}-\mu_t)}\leq a_t,~\frac{\mu_s(\widehat{g_s}-\mu_b)}{\mu_b(\widehat{g_s}-\mu_s)}\leq a_s.
  \end{split}
\end{equation}
The final feasible sets $Q_c$ for each $a_c$ will be correspondingly the intersection set of \eqref{Equ:Constraints} and \eqref{Equ:Constraints_feasible}\footnote{An implicitly constraint $\mu_s\leq\mu_b\leq \mu_t$ always holds according to the semantic prior discussed in the previous section.}. If we ignore the coupled variables in the feasible set to obtain the approximate solutions for the sake of simplicity and efficiency, we can optimize each variable with the following iterative scheme, where all shape features $a_c$ can be empirically initialized as 1.
\begin{equation}\label{Equ:Para_iterative}
\begin{split}
  a_t^{(t+1)}&\gets\arg\min h(a_t|N_t,\mathbf{g}_t)~\mathrm{s.t.}~a_t\in Q_a^{(t)},\\
  a_b^{(t+1)}&\gets\arg\min h(a_b|N_b,\mathbf{g}_b)~\mathrm{s.t.}~a_b\in Q_b^{(t+1)},\\
    a_s^{(t+1)}&\gets\arg\min h(a_s|N_s,\mathbf{g}_s)~\mathrm{s.t.}~a_s\in Q_s^{(t+1)}.\\
\end{split}
\end{equation}
$h(a|N,\mathbf{g})=N\log(\Gamma(a))+Na\log(\mu/a)+aN-a\sum_{i=1}^{N}\log(|g_i|+\epsilon)$ is a simple univariate differentiable function. The solution to any a problem in \eqref{Equ:Para_iterative} will exist either on the boundary of the feasible set or the root of the derivative function $h'(a|N,\mathbf{g})=0$ as $h'(a|N,\mathbf{g})$ is a monotone function. However, no closed form solution of the root can be derived so that we will instead use the following approximation\footnote{http://research.microsoft.com/en-us/um/people/minka/papers/minka-gamma.pdf}. The iteration stops when the variation of objective function \eqref{Opt:Paraac} is below a threshold.
\begin{equation}\label{Sol:ac_approx}\small
  a^*=\frac{3-s+\sqrt{(3-s)^2+24s}}{12s},~s=\log(\mu)-\frac{1}{N}\sum_{j=1}^{N}\log(|g_j|+\epsilon).
\end{equation}
The global optimization stops when the iteration number reaches the maximum or the variation of $\mathbf{G}$ between two iterations is below a predefined threshold and it is summarized in following \textbf{Algorithm} \ref{alg:TO-SARImaging}.
\begin{algorithm}
  \caption{Semantic information guided target oriented SAR image formation}
  \label{alg:TO-SARImaging}
  \begin{algorithmic}
  \REQUIRE Sampled phase history $\mathbf{\widetilde{r}}$, Measurement operator $\widetilde{\Phi}$, Parameters $\beta$, $\lambda_0$, Maximum iteration $T$.
  \ENSURE target enhanced image: $\mathbf{G}$.
  \STATE \textbf{initialization: }$\mathbf{G}^{(0)}=\mathbf{\widetilde{\Phi}}^{*}(\mathbf{\widetilde{r}})$, $a_c^{(0)}=1$, weight matrix: $\mathbf{W}^{(0)}=\mathbf{I}$, $k=1$.
    \WHILE{$k\leq T$ or not converge}
  \IF{$k=1$}
  \STATE Image $\mathbf{G}^{(k)}$ recovery based on with weighted $\ell_1$ minimization.
  \ELSE
  \STATE Image $\mathbf{G}^{(k)}$ recovery with IRW-$\ell_1$ minimization starting with $\mathbf{W}^{(k)}$.
  \ENDIF
     \STATE Semantic label inference $\mathbf{Y}^{(k)}$ according to current $\mathbf{G}^{(k)}$.
    \STATE Refine $\mathbf{Y}^{(k)}$ according to the semantic size prior.
  \STATE Update the features $\Theta^{(k)}$.
  \STATE Update the weight matrix $\mathbf{W}^{(k)}$ according to current $\mathbf{G}^{(k)}$, $\mathbf{Y}^{(k)}$ and $\Theta^{(k)}$.
  \STATE $k\gets k+1$.
  \ENDWHILE
  \end{algorithmic}
\end{algorithm}
\subsection{Convergence and Complexity Analysis}\label{Subsec:convergence}
\par
\begin{figure}
  \centering
  \subfigure[]{\includegraphics[width=0.15\textwidth]{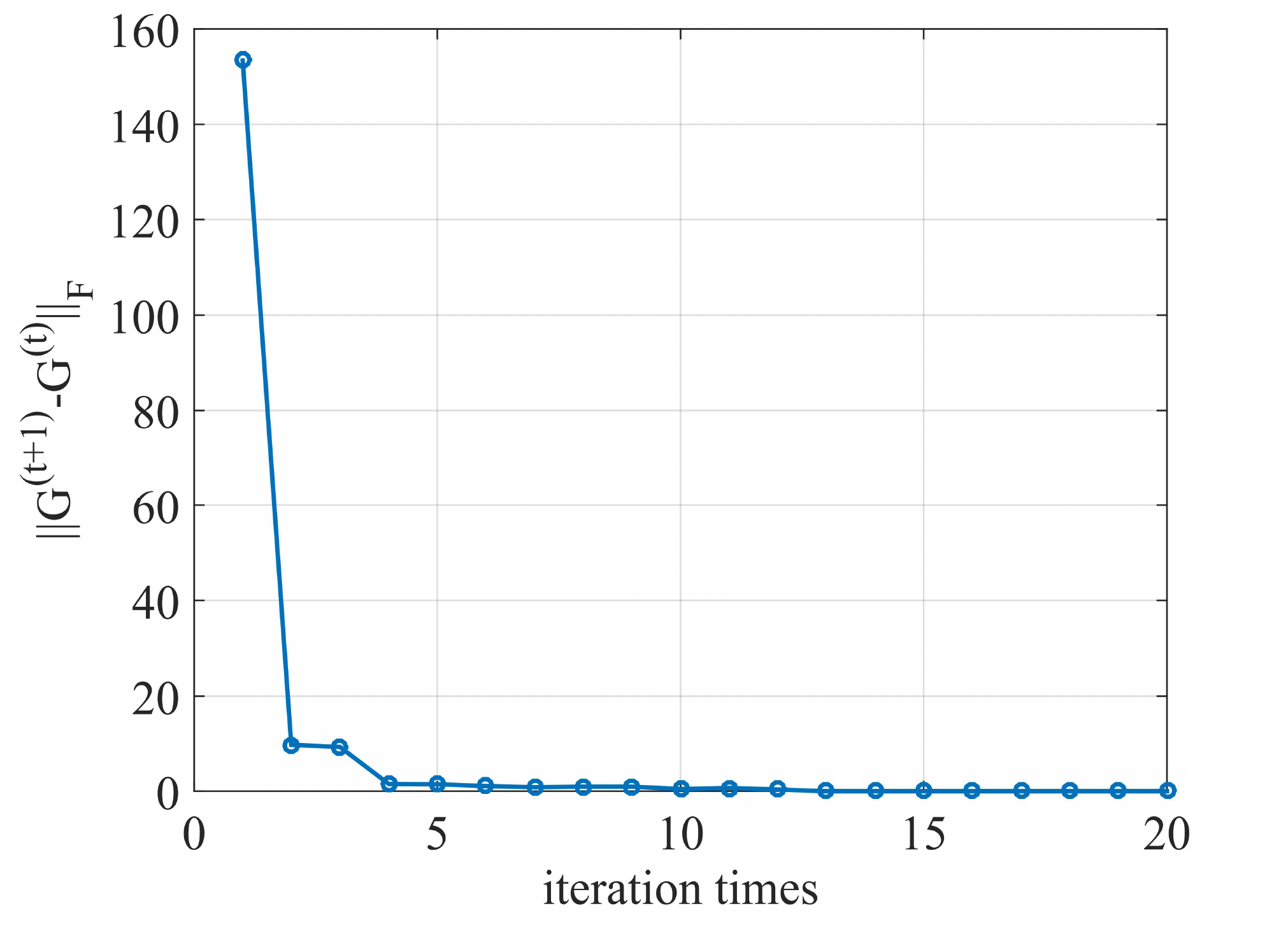}
 \label{fig_ConveMM}}
  \subfigure[]{\includegraphics[width=0.15\textwidth]{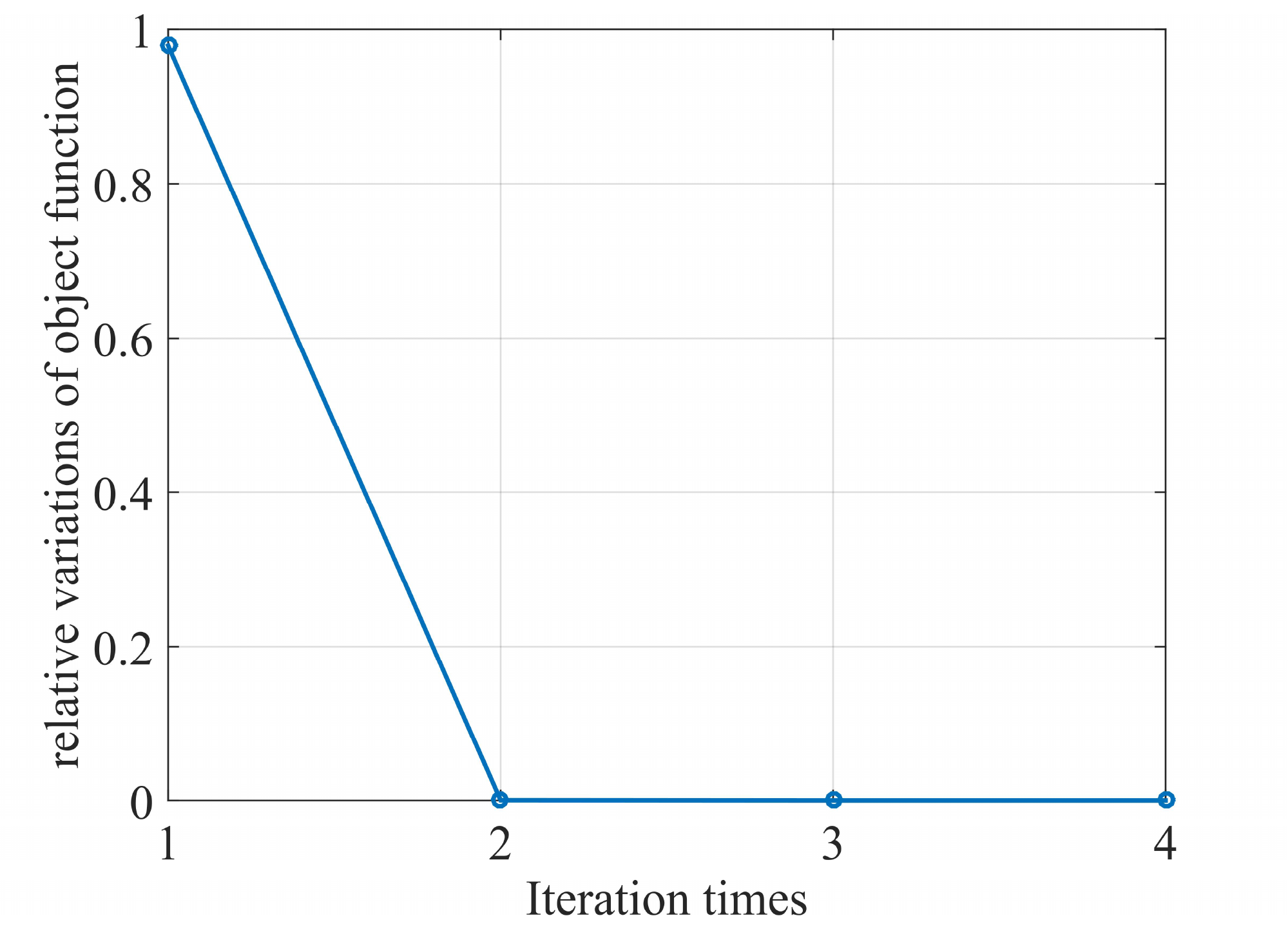}
 \label{fig_ConvFista}}
   \subfigure[]{\includegraphics[width=0.15\textwidth]{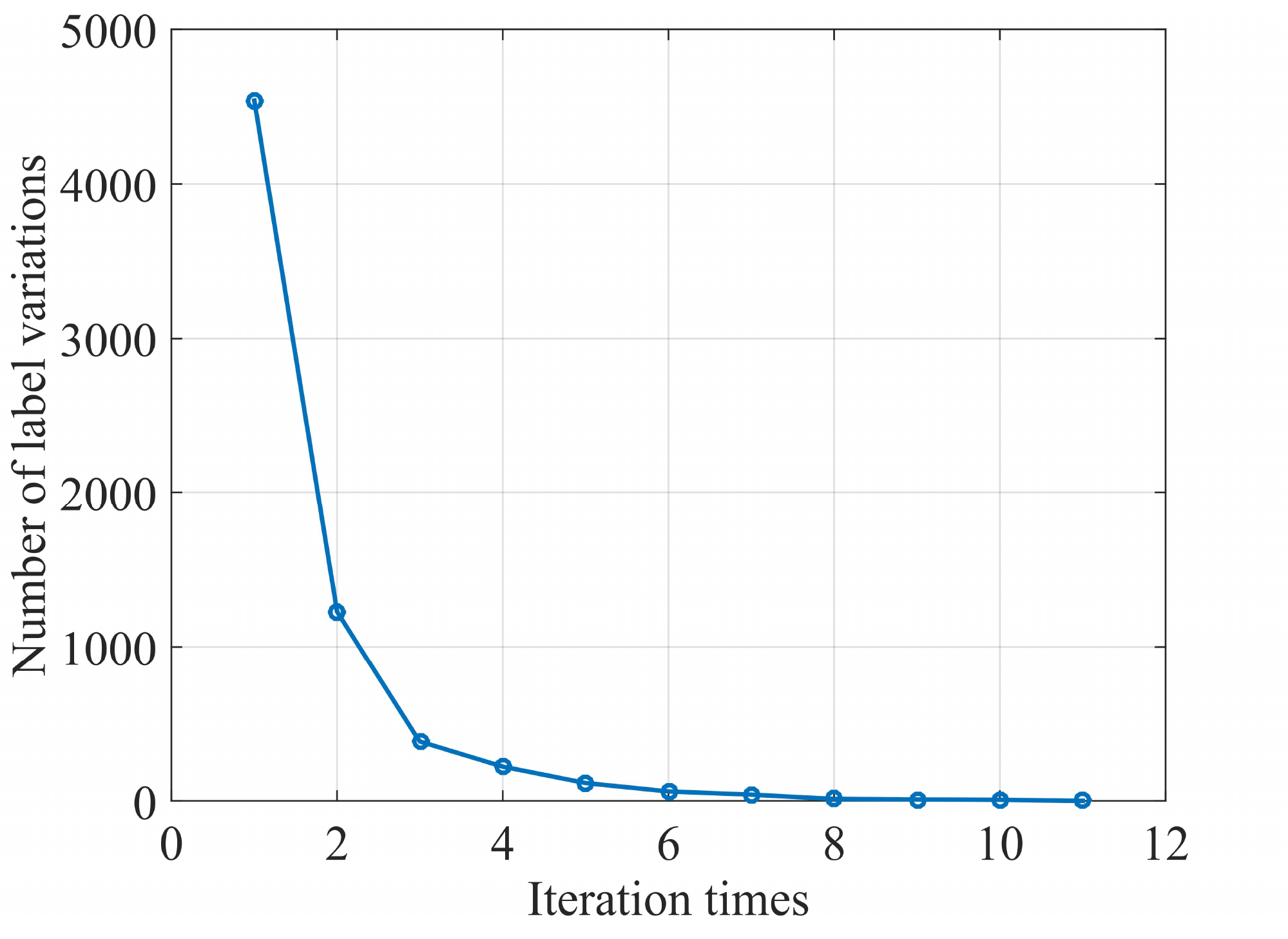}
 \label{fig_ConvICM}}
  \caption{Convergence curves of (a) IRW-$\ell_1$ minimization with MM algorithm, (b) weighted $\ell_1$ minimization with FISTA algorithm, (c) label inference with ICM.}\label{Fig_illustratetargetshadow}
\end{figure}

 Since the exactly theoretical convergence proof for the global \textbf{Algorithm} \ref{alg:TO-SARImaging} is relatively difficult, we only present the analysis and experimental validation on each algorithm in different phases, namely MM scheme for IRW-$\ell_1$ minimization, FISTA for weighted $\ell_1$ minimization and label inference with ICM and we choose the corresponding results from the first global iteration for illustration. It has been discussed in Sec. \ref{subsec:IRWL1} that the MM algorithm aims to solve a concave-convex optimization \eqref{Equ:IterativeSAR} by exploiting and minimizing a surrogate function. This algorithm can iteratively decrease the objective function by \eqref{Equ:ConverMM} and generally converge to a local minimum \cite{Candes2008}. We plot the variances $\|\mathbf{G}^{(t+1)}-\mathbf{G}^{(t)}\|_\mathrm{F}$ vs. MM iteration times in Fig. \ref{fig_ConveMM} and we can see that the algorithm converges rapidly within 5 iterations to a stationary point. FISTA is a widely exploited solver for weighted $\ell_1$ minimization that is benefit from its simplicity for large scaled problem and fast convergence rate of $\mathcal{O}(1/t^2)$ \cite{beck2009fast}, where $t$ stands for the iteration in this algorithm. Since this subproblem is a convex one, it can converge to a global solution ignoring the initial point.  We plot the variations of objective function vs. iteration times in Fig. \ref{fig_ConvFista} and the algorithm stops only within 4 iterations. Finally, the greedy ICM algorithm is adopted for label inference, whose solution converges to a local minimal \cite{Besag1986}. The number of variational labels vs. ICM iteration times is plotted in Fig. \ref{fig_ConvICM}, where the algorithm also appears a rapid convergence.
\par Considering the computational complexity, the main computations focus on solving weighted $\ell_1$ minimization with FISTA since label inference and features update are the both simple univariate and unidimensional optimizations. Firstly, FISTA essentially involves three main operations, namely $\widetilde{\Phi}^{*}(\mathbf{r})$, $\widetilde{\Phi}^{*}\widetilde{\Phi}(\mathbf{X})$ and $\mathcal{S}_\mathbf{\lambda,W}(\mathbf{X})$. $\widetilde{\Phi}^{*}\widetilde{\Phi}(\mathbf{X})=\mathbf{A}^{*}(\Phi^{*}\Phi)\mathbf{A}(\mathbf{X})$ in fact consists of three consecutive operations. $\mathbf{A}$ and $\mathbf{A}^*$ can be realised by performing 2D-DFT and 2D-iDFT with very cheap computational complexity. $\Phi^{*}\Phi$ sets those unsampled entries to be 0 via simply binary masking. $\widetilde{\Phi}^{*}({\mathbf{r}})=\mathbf{A}^{*}\Phi^{*}(\mathbf{r})$ involves two consecutive operations, in which $\mathbf{{r}}$ is firstly interpolated with zero padding and following a 2D-iDFT operation. For $\mathbf{X}\in\mathbb{C}^{N\times K}$, the element-wise shrinkage operator $\mathcal{S}_\mathbf{\lambda,W}(\mathbf{X})$ only needs $\mathcal{O}(NK)$ operations.
\section{Experiments and Discussion}\label{Sec:Experiments}
\begin{figure*}
  \centering
  \includegraphics[width=0.8\textwidth]{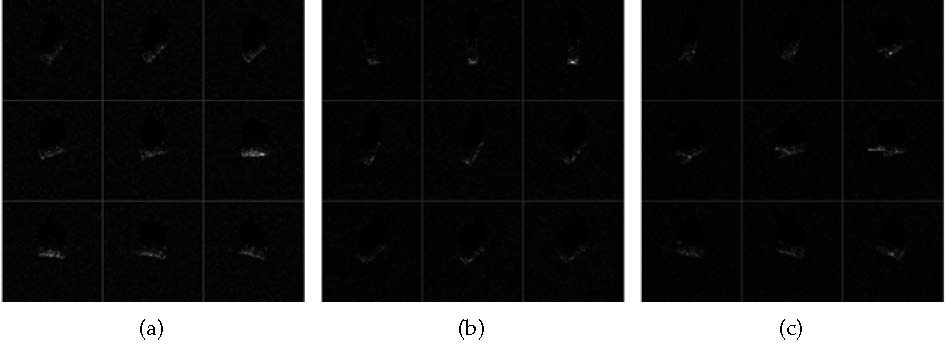}\\
  \caption{Illustration the magnitude images of some samples from MSTAR database . (a) BMP2 tanks. (b) BTR70 armored personnel carriers. (c). T72 tanks.}\label{Fig_SampleImage}
\end{figure*}

In this section, we will validate the performance of our target oriented SAR image formation framework with a series of experiments on the public MSTAR target database \cite{Hummel2000}. This database collected by the Sandia National Laboratories Twin Otter spotlight mode SAR platform provides the complex SAR images of various military vehicles and we exploit three types of the target taken at $17^\circ$ depression angle for testing, namely BMP2 tanks (BMP2-SN9566), T72 tanks (T72-SN132) and BTR70 armored personnel carriers (BTR70-SNC71). Some $128\times 128$ sized magnitude images are exhibited in Figs. \ref{Fig_SampleImage} and some detailed information of these targets are summarised in Table \ref{SizeInformation}.

\begin{table}\small
  \centering
  \caption{Some Information of the Exploited Targets for Testing}\label{SizeInformation}
  \begin{tabular}{c|c|c|c}
  \hline
  \hline
  Target Type&Sequence & Target Size (m)& Resolution (m)\\
  \hline
  BMP2&sn-9566&$6.71\times 3.09\times 2.06 $&$0.3\times 0.3$\\
  BTR70&sn-c71&$7.54\times 2.80\times 2.32 $&$0.3\times 0.3$\\
  T72&sn-132 &$6.95\times 3.59 \times 2.22 $&$0.3\times 0.3$\\
  \hline
  \end{tabular}

\end{table}
\par Following previous works \cite{cetin2001feature}\cite{cetin2003feature}, the sampled phase history data can be simulated from the complex MSTAR image as following. Firstly, each complex SAR image is transformed into the spatial-frequency domain by performing the 2D-DFT operation, then we remove the surrounding 28 pixels-width zero padding data to obtain a $100\times 100$ 2D array. From the file headers, a 35dB Taylor window has been added on the phase history so that the fully sampled raw phase history $\mathbf{r}$ will be obtained by dividing this array with the 2D Taylor window with $\overline{n}=4$. The undersampled phase history vector $\widetilde{\mathbf{r}}$ will be subsequently simulated as $\widetilde{\mathbf{r}}=\Phi \mathbf{r}$.
\par In order to demonstrate the effectiveness and superiority of our high level semantic-specific regularization function for target oriented SAR image formation, two typical sparsity-driven functions for low level feature enhancement \cite{Patel2010}\cite{cetin2001feature} will be primarily compared, including $\ell_1$ regularization for point feature enhancement (Poi-Imaging) and TV regularizer for region smoothness enhancement (Reg-Imaging). Additionally, to show the performance of our weights (semantic features) in IRW-$\ell_1$, we also compare the standard IRW-$\ell_1$ framework for SAR imaging \cite{Ma2015}\cite{Candes2008}, in which weights are determined by the magnitude of each pixel without considering their semantic information. Moreover, the image reconstructed from the conventional polar formatting algorithm (PF-Imaging) is also shown compared, which is actually the initial solution used in above imaging algorithms. In addition to visual evaluation, some quantitative criterions will be computed to evaluate the performance of target enhancement as well as background suppression, including 1) peak target-to-clutter ratio (PTCR): $20\log_{10}\frac{N\widehat{g_t}}{\sum_{i=1}^N{g_i}},~\forall y_i\neq ``t"$.  and 2) average target intensity: $\chi_t=\|\mathbf{g}_t\|_2^2/N_t$. To compute these criterions, the manually segmentation result of each target image will be served as the ground truth.
\subsection{Parameter Analysis}\label{Subsec:paraanalysis}
In the proposed framework, two free parameters needs to be tuned in advance, including $\lambda_0$ and $\beta$. We will conduct two groups of experiments to validate their performances, where only $50\%$ phase history data are exploited in random for imaging.
\par
\begin{figure*}
  \centering
  \includegraphics[width=0.8\textwidth]{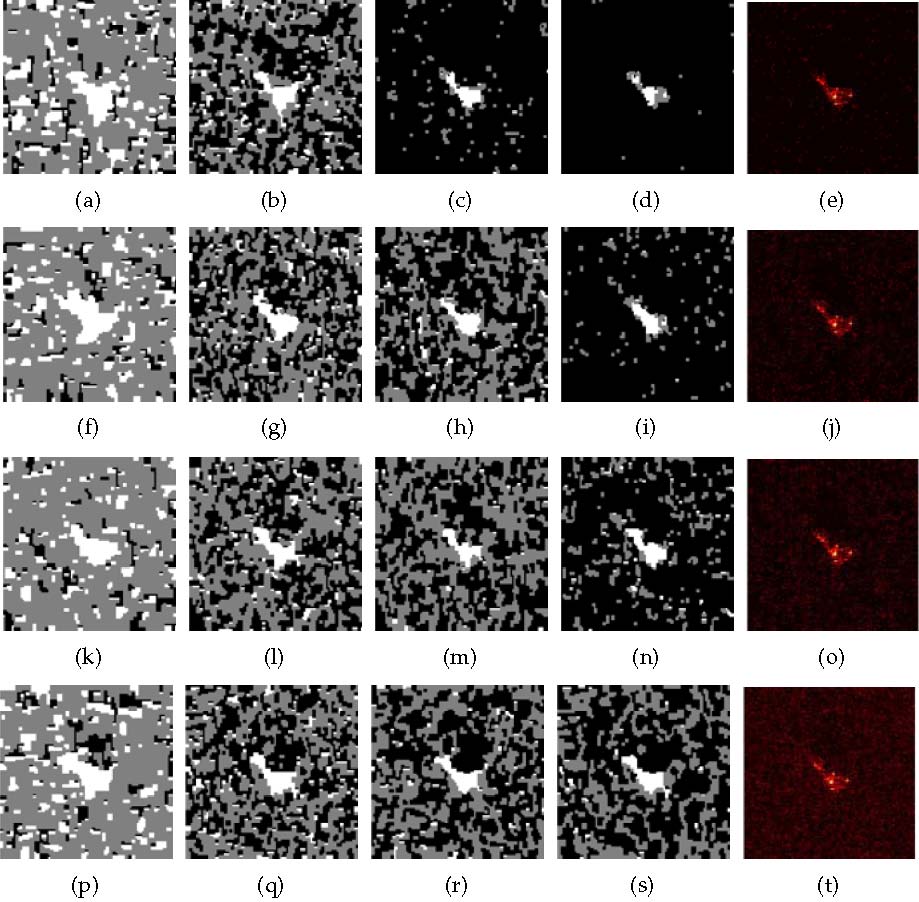}\\
  \caption{Label maps  in different iterations and reconstructed image with different values of $\lambda_0$.  Row: (a)-(d) label map with $\lambda_0=10$. (f)-(i) label map with $\lambda_0=30$. (k)-(n) label map with $\lambda_0=50$. (p)-(s) label map with $\lambda_0=70$.  Column: (a)-(p) label map in the 1st iteration. (b)-(q) label map in the 2nd iteration. (c)-(r) label map in the 3rd iteration. (d)-(s) label map in the 4th iteration. (e)-(t) reconstructed image in the fifth iteration.}\label{Fig_lambda}
\end{figure*}

 Firstly, $\lambda_0$ controls the increasing rate of the regularization parameter $\lambda$, which is introduced to progressively suppress the background pixels and avoid target missing. Nevertheless, larger value of $\lambda_0$ will generally not only result in a slower increase of $\lambda$ but will also decelerate the algorithm convergence. Therefore, an appropriate value of $\lambda_0$ is required to achieve a balance between the accuracy of semantic label inference and convergence. To this end, we vary $\lambda_0$ from 10 to 70 with 20 interval and display some of the intermediate results of the inferred label maps in the first four iterations and the reconstructed images in the fifth iteration in hot color map in Figs. \ref{Fig_lambda}, where the white, black and grey color stand for the target, shadow and background class, respectively.

\par Observing the estimated label maps Figs. 7(a)-(d), we essentially only concern the completeness of the target region rather than background and shadow\footnote{This is different from the semantic segmentation task which requires an exactly accuracy for every class.}. In these maps, as our analysis in Sec. \ref{Subsec:Analysis}, some target points are obvious misclassified with the proceeding of iteration, leading to a target missing label map in Fig. 7(d). Such problem is effectively relieved by an increase of $\lambda_0$ and the corresponding label maps in Fig 7(n) and 7(s) will cover almost all target points and preserve the target shape perfectly. On the contrary, inspecting the last column of Figs. \ref{Fig_lambda}, the most background pixels have already been suppressed with rather low energy in the 5th iteration shown in Fig. 7(e) while the clutters in Fig. 7(t) are obvious with much larger energies. This phenomenon indicates a deceleration in the algorithm convergence. Therefore, we will select $\lambda_0=50$ in the following experiments to reach a tradeoff.
\par
\begin{figure*}
  \centering
  \includegraphics[width=0.8\textwidth]{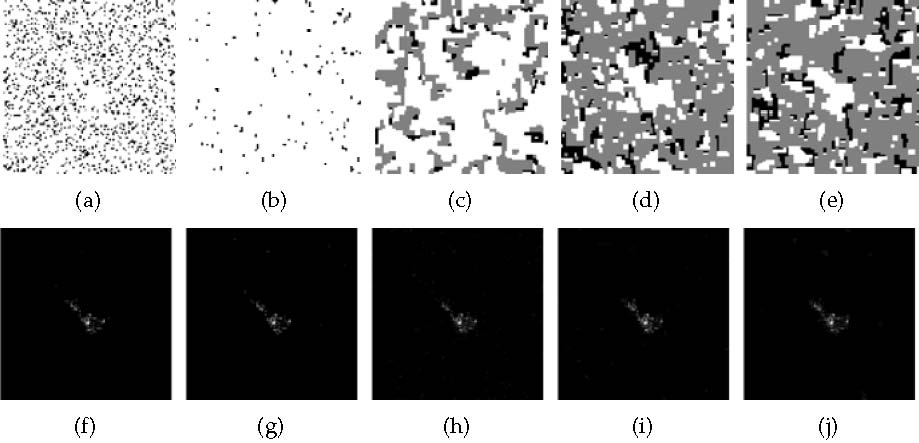}\\
  \caption{Initial inferred label maps and the final recovered target images with different values of $\beta$. First row: label maps. Second row: recovered target images. (a)(f) $\beta=0.01$. (b)(g) $\beta=0.1$. (c)(h) $\beta=0.5$. (d)(i) $\beta=1$. (e)(j) $\beta=5$.}\label{Fig_beta}
\end{figure*}

 Secondly, $\beta$ is the parameter of label prior distribution, which provides a tradeoff between likelihood and  prior-inducing function in Eq. \eqref{Opt:SemanticInference}. Two types of priors are actually involved in the function, i.e., local label consistency and co-occurrence rules. On one hand, if $\beta$ is selected as a larger value, the inferred label map will intuitively be more consistent within each cluster group, and vice versa. On the other hand, a smaller value of $\beta$ may lead to an inaccurate or false inference due to the semantic gap. To verify these viewpoints, we test $\beta$ with values of $\{0.01,0.1,0.5,1,5\}$ and show the corresponding inferred label map (without semantic refinement) in the first iteration in Figs. \ref{Fig_beta}.

\par In Fig. 8(a), the label map appears many isolated points in the case of $\beta=0.01$. When the value of $\beta$ increases, such isolated points gradually disappear and more continuous regions emerge in Figs. 8(b) and (c). But the inferred labels are still incorrect for the most background and shadow points. When $\beta=1$, the shape of the target can be outlined and some shadow points of the target emerge. It is also worth noting that although $\beta$ heavily influence the initial label inference, our proposed framework can also make a refinement in the subsequent iteration as discussed in Figs. \ref{Fig_lambda}. Therefore, we can always obtain a desired target enhanced image with different choices of $\beta$, which are also correspondingly illustrated in the second row in Figs. \ref{Fig_beta}. Following consideration of the simplicity, we will set $\beta=1$ in the following experiments.
\subsection{Framework Validation}\label{subsec:ExperimentFrameworkValidation}
\par
\begin{figure*}
  \centering
  \includegraphics[width=0.8\textwidth]{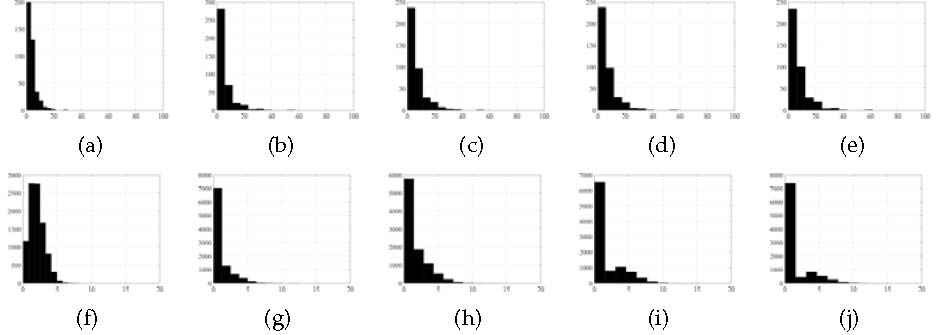}\\
  \caption{Illustration of the histograms of the target and background pixels in different iterations. (a)-(e) target histograms in 1st to 5th iteration. (f)-(g) background histograms in 1st to 5th iteration.}\label{Fig_HIstVari}
\end{figure*}

 In previous experiments, we have empirically established two important parameters in the framework and show that the label inference can be gradually refined with the iterative proceeding. In this part, the core issue of the proposed framework will be further validated, i.e., whether the framework can produce a target enhanced image in progress. To demonstrate this issue, the histograms in different iterations of target pixels and background clutters including the shadow points are illustrated in Figs. \ref{Fig_HIstVari}, respectively. According to the target histograms, the peak value is obviously enhanced from almost 30 to 60. After a careful inspection of the second row of histograms of the background clutters, more and more background pixels gradually tend to zero along with the iteration. Comparing Fig. 9(f) with Fig. 9(j), the number of background pixels approximating 0 is from less than 3000 to over 7000 while the target histograms almost keep unchanged. Therefore, the average magnitude of the background pixels is greatly decreased so as to yield an increase of PTCR. The experimental results clearly demonstrate the effectiveness of proposed iterative framework for target enhancement.

\par
\begin{figure*}
  \centering
  \includegraphics[width=0.8\textwidth]{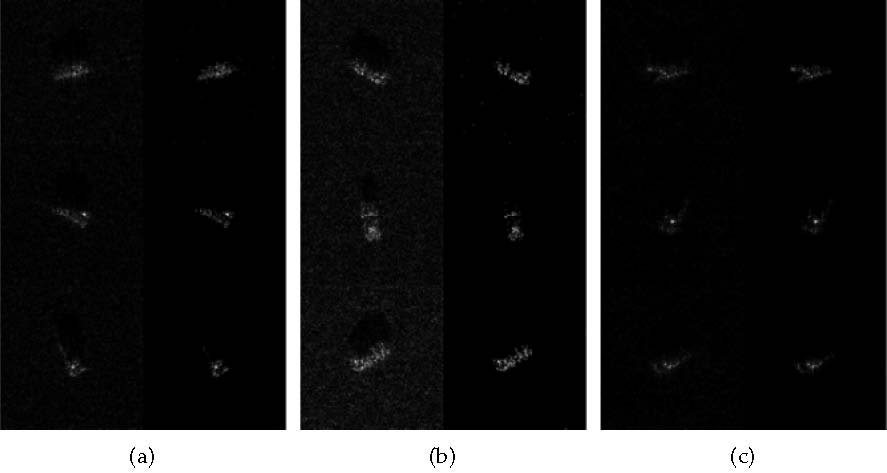}\\
  \caption{Illustration of the reconstructed images for different types of targets. left column: the primary image. right column: Tar-Imaging result. (a) BMP2. (b) BTR70. (c) T72.}\label{Fig_DifferentTarget}
\end{figure*}

 Next we will investigate the robustness of the proposed target oriented framework to different types of targets. Following the same parameters setting without further fine tuning, different types of the targets are tested for reconstruction and some of the results are shown in Figs. \ref{Fig_DifferentTarget}, where the corresponding primary images are also presented for comparison. It is obviously observed from the results that our framework is robust to different types of target as it can always generate an apparent target enhanced SAR image.

\subsection{Framework Comparison with Different Sampling Schemes}
\par
\begin{figure*}
  \centering
  \includegraphics[width=0.8\textwidth]{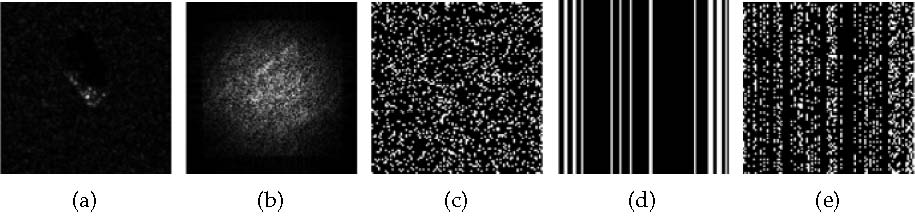}\\
  \caption{Illustration of the simulated phase history data (without removing the zero padding entries) and three types of binary masks. (a) primary sample image. (b) phase history data without removing zero-padding. (c) Mask-1. (d) Mask-2. (e) Mask-3.}\label{Fig_mask}
\end{figure*}

 In the final experiments, we are to compare the proposed target oriented framework with other SAR imaging algorithms in various undersampling situations, where the so called undersampling rate $\eta$ is defined as the ratio between the number of entries in $\widetilde{\mathbf{r}}$ and that of $\mathbf{r}$. In our experiments, three types of undersampling schemes are simulated by the binary masks shown in Figs \ref{Fig_mask}. From the figures, a pure global 2D random sampling from a Cartesian grid is instantiated by Mask-1 in Fig. 11(c). Nevertheless, this type of sampling scheme is ineffective in practical imaging situation because the complete phase history data from all viewing angles are still required. To relieve the burden on the requirement of a large amount of viewing angles for a high cross range resolution, a more suitable scheme is simulated as Mask-2 in Fig. 11(d) by which we can only randomly sample the received data from a few observation angles. In this case, the undersampling rate is simply computed by the ratio between the number of the sampled viewing angles and that of the complete viewing angles denoted by $\eta=\eta_c$. Moreover, the sampling burden of a SAR platform mostly comes from the range direction to produce a high range resolution image, which conventionally requires a Nyquist sampling speed. To relieve this burden, undersampling in range direction is also taken into consideration as Mask-3 in Fig. 11(e), where we not only randomly select a few viewing angles but also randomly pick samples of the corresponding phase history. If the undersampling rate in range direction for all viewing angles is the same as $\eta_r$, the total undersampling rate will be $\eta=\eta_r\eta_c$. Therefore, Mask-2 can be viewed as a special type of Mask-3 with $\eta_r=1$. {In the theory of compressive sampling, one of the sufficient conditions for exactly recovery is K-order restricted isometry property (RIP) \cite{candes2006robust}, which provides a theoretical guarantee for the uniqueness recovery from an underdetermined problem with sparse regularization.} In our case, we formulate our target oriented imaging as a MAP and thus it is essentially not an exactly recovery problem so that it is unnecessary to require such a condition on the measurement matrix. Nevertheless, the equivalent measurement matrix in our framework is the partial Fourier matrix whose RIP has been validated in many previous works \cite{Patel2010}\cite{Becker2011}.

\subsubsection{Sampling with Mask-1} Firstly, we exploit Mask-1 for SAR imaging with different algorithms in which the undersampling rate $\eta$ is chosen as $15\%$, $20\%$ and $50\%$. The resulted sample images are illustrated in following Figs \ref{Fig_GlobalSampling} and the average quantitative evaluations computed from all tested images are listed in Table. \ref{Tab:GlobalSampling}.

\begin{figure*}
  \centering
  \includegraphics[width=0.8\textwidth]{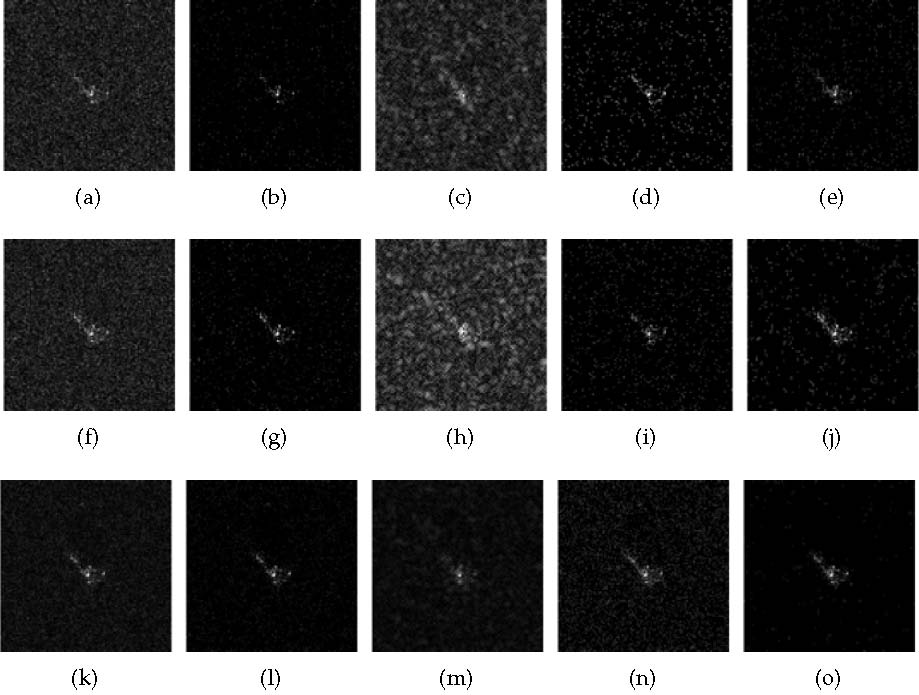}\\
  \caption{SAR image formation from undersampled phase history data with Mask-1 according to different imaging algorithms. Row: (a)-(e) $\eta=15\%$. (f)-(j) $\eta=20\%$. (k)-(o) $\eta=50\%$. Column: (a)-(k) PF-Imaging. (b)-(l) Poi-Imaging. (c)-(m) Reg-Imaging. (d)-(n) IRW-$\ell_1$. (e)-(o) Tar-Imaging.}\label{Fig_GlobalSampling}
\end{figure*}

\begin{table*}\small
\centering
\caption{Quantitative Evaluation for SAR image Formation from Partially Observed Phase History Data with MASK-1.}\label{Tab:GlobalSampling}
\begin{tabular}{|c|c|c|c|c|c|c|}
\hline
BTR& \multicolumn{2}{c|}{$15\%$} & \multicolumn{2}{c|}{$20\%$} & \multicolumn{2}{c|}{$50\%$}\tabularnewline
\hline
Algorithm & PTCR & $\overline{\chi_{t}}$ & PTCR  & $\overline{\chi_{t}}$ & PTCR  & $\overline{\chi_{t}}$\tabularnewline
\hline
PF-Imaging &17.0008   &7.7981  &  18.0431 &12.7282   &  22.4897 &  66.1276    \tabularnewline
\hline
Poi-Imaging &41.0656   & 91.4789 & 39.5539   & 126.0911  & 32.2728  &   98.9543      \tabularnewline
\hline
Reg-Imaging & 14.6138  &1.1394  & 15.8597 &   2.1761  &   23.6771  &     47.6311      \tabularnewline
\hline
IRW-$\ell_{1}$ &36.6725  & 122.8220 &35.6254  &  156.3631  & 31.4178   &   190.5878     \tabularnewline
\hline
Tar-Imaging &\textbf{43.9012}  &\textbf{178.5304}  &\textbf{52.9407} &  \textbf{203.5214} &   \textbf{67.7273}
 &     \textbf{212.9420  } \tabularnewline
\hline
\hline
T72& \multicolumn{2}{c|}{$15\%$} & \multicolumn{2}{c|}{$20\%$} & \multicolumn{2}{c|}{$50\%$}\tabularnewline
\hline
Algorithm & PTCR & $\overline{\chi_{t}}$ & PTCR  & $\overline{\chi_{t}}$ & PTCR  & $\overline{\chi_{t}}$\tabularnewline
\hline
PF-Imaging &13.0623   & 7.5820 & 13.5501   &  11.6905 & 15.6163  &   54.9055     \tabularnewline
\hline
Poi-Imaging & 27.4130 & 81.6110  & 26.0462   &   95.1180   &18.6934  &   57.4675      \tabularnewline
\hline
Reg-Imaging &  11.7601   & 1.2717  & 12.5823 &  1.9718  &  16.4631&     19.0626    \tabularnewline
\hline
IRW-$\ell_{1}$ & 25.6452   & 106.5971  & 24.2680 & 121.0326  &  20.2626 &    142.5146     \tabularnewline
\hline
Tar-Imaging & \textbf{30.0898} & \textbf{140.9991}& \textbf{31.3511}& \textbf{157.0146} & \textbf{35.6465} &     \textbf{164.0307}   \tabularnewline
\hline
%
\hline
BMP2& \multicolumn{2}{c|}{$15\%$} & \multicolumn{2}{c|}{$20\%$} & \multicolumn{2}{c|}{$50\%$}\tabularnewline
\hline
Algorithm & PTCR & $\overline{\chi_{t}}$ & PTCR  & $\overline{\chi_{t}}$ & PTCR  & $\overline{\chi_{t}}$\tabularnewline
\hline
PF-Imaging & 18.3315 & 12.5190 &  19.4606  &  20.8500  &  23.9939  &    109.2194   \tabularnewline
\hline
Poi-Imaging & 42.9390 &187.5272  & 41.4655  &  220.1853  & 32.1583  &   133.4031    \tabularnewline
\hline
Reg-Imaging & 15.7512 &  1.7024 & 17.3495  & 3.3094  & 26.6544 &  83.3213     \tabularnewline
\hline
IRW-$\ell_{1}$ &38.3855  &  220.9719 & 37.4246 &  256.4996 &  32.8173 &    327.6412      \tabularnewline
\hline
Tar-Imaging & \textbf{48.2299} & \textbf{324.4334} &  \textbf{57.7635} &  \textbf{337.1599} & \textbf{71.2109} &  \textbf{ 343.3820}    \tabularnewline
\hline
\end{tabular}
\end{table*}
\par Observing Figs. \ref{Fig_GlobalSampling} row by row at the first sight, the visual qualities of all the reconstructed images are progressively improved with the increase of $\eta$. We also conduct the experiment in the case of $\eta<10\%$, but the resulted target images of all algorithms will be hardly recognized so that we will not display those the failing reconstruction results for the sake of space limitation. Among all Figs. \ref{Fig_GlobalSampling}, the target images of Reg-Imaging are the worst from the visual perspective as it is already difficult to manually distinguish the target from these results, let alone automatic target recognition by a computer. This is mainly due to the fact that TV regularizer will concentrate on preserving the edges in the image. When the variance of the background clutters is large, many clutters will be falsely regarded as the edges so as to degrade the image. Looking up the Table \ref{Tab:GlobalSampling}, the quantitative evaluations in terms of average PTCR and $\overline{\chi_t}$ are also the worst for all types of the targets and $\eta$. We can see that $\overline{\chi_t}$ and PTCR are less than 5 and 20dB in the case of a small amount of $\eta$. Until $\eta$ is increased to $50\%$, its PTCR can achieve the slightly higher values than that of PF-Imaging, while $\chi_t$ are still the worst. Concerning the results of PF-Imaging in Figs. 12(a),(f) and (k), the target in Fig. 13(a) becomes rather blurry and merges in the background clutters, respectively. By checking the quantitative evaluations from Table \ref{Tab:GlobalSampling}, the values are only better than Reg-Imaging algorithm. The rest three algorithms including Poi-Imaging, IRW-$\ell_1$ and Tar-Imaging all outperform PF-Imaging and Reg-Imaging in both visual and quantitative evaluations as following discussion. Firstly, the target images obtained from Poi-Imaging, IRW-$\ell_1$ and Tar-Imaging are more clear than that of PF-Imaging and Reg-Imaging, especially the resolution of target scatter points is greatly improved, which is actually benefit from the $\ell_1$ norm for point feature enhancement. Moreover, these three algorithms gain a remarkable better quantitative indexes. Comparing these three algorithms, our Tar-Imaging and IRW-$\ell_1$ can obtain a better visual image than Poi-Imaging when the sampling rate is only $15\%$. We can see from Figs. 12(b),(d) and (e), the target in the latter two images is more complete in shape and profile. This result verifies the superiority of the iterative reweighted framework for reducing the required measurements. Nevertheless, it can be also observed from Fig. 12(d) that many background clutters are also enhanced while those in our 12(e) are suppressed due to the consideration of their semantic labels. We can further validate this issue from the quantitative performances in Table \ref{Tab:GlobalSampling}, in which our algorithm achieves a significant improvement on both PTCR and $\chi_t$. Specifically, our PTCR in different cases are almost more than twice of others, which clearly show the superiority of target enhancement and background clutter suppression.
\subsubsection{Sampling with Mask-2 and Mask-3} Next, we will compare all SAR imaging algorithms with Mask-2 and Mask-3 together since Mask-2 is a special type of Mask-3 with $\eta_r=1$. In the experiments, seven pairs of sampling schemes are exploited, including $\eta_c=20\%,~\eta_r=1$, $\eta_c=25\%,~\eta_r=1$ and $\eta_c=50\%,~\eta_r=1$ for Mask-2 and $\eta_c=70\%,~ \eta_r=30\%$, $\eta_c=50\%,~\eta_r=30\%$, $\eta_c=50\%,~\eta_r=50\%$ and $\eta_c=50\%,~\eta_r=70\%$ for Mask-3.   The imaging results of the sample image are illustrated in following Figs. \ref{Fig_CrossSampling} and \ref{Fig_CrossandrangeSampling}, respectively  and the average quantitative evaluations computed from all tested images are listed in Table \ref{Tab:crossrangeandrangeSampling}.
\begin{figure*}
  \centering
  \includegraphics[width=0.9\textwidth]{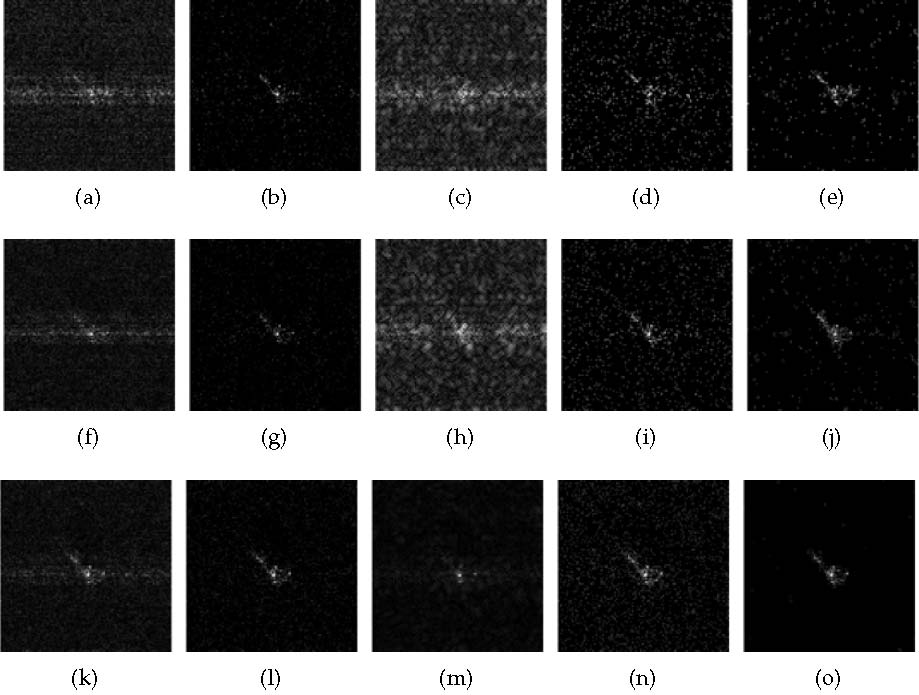}\\
  \caption{SAR image formation from undersampled phase history data with MASK-2 according to different imaging algorithms. Row: (a)-(e) $\eta_c=20\%$. (f)-(j) $\eta_c=25\%$. (k)-(o) $\eta_c=50\%$. Column: (a)-(k) PF-Imaging. (b)-(l) Poi-Imaging. (c)-(m) Reg-Imaging. (d)-(n) IRW-$\ell_1$. (e)-(o) Tar-Imaging.}\label{Fig_CrossSampling}
\end{figure*}
\begin{figure*}
  \centering
  \includegraphics[width=0.9\textwidth]{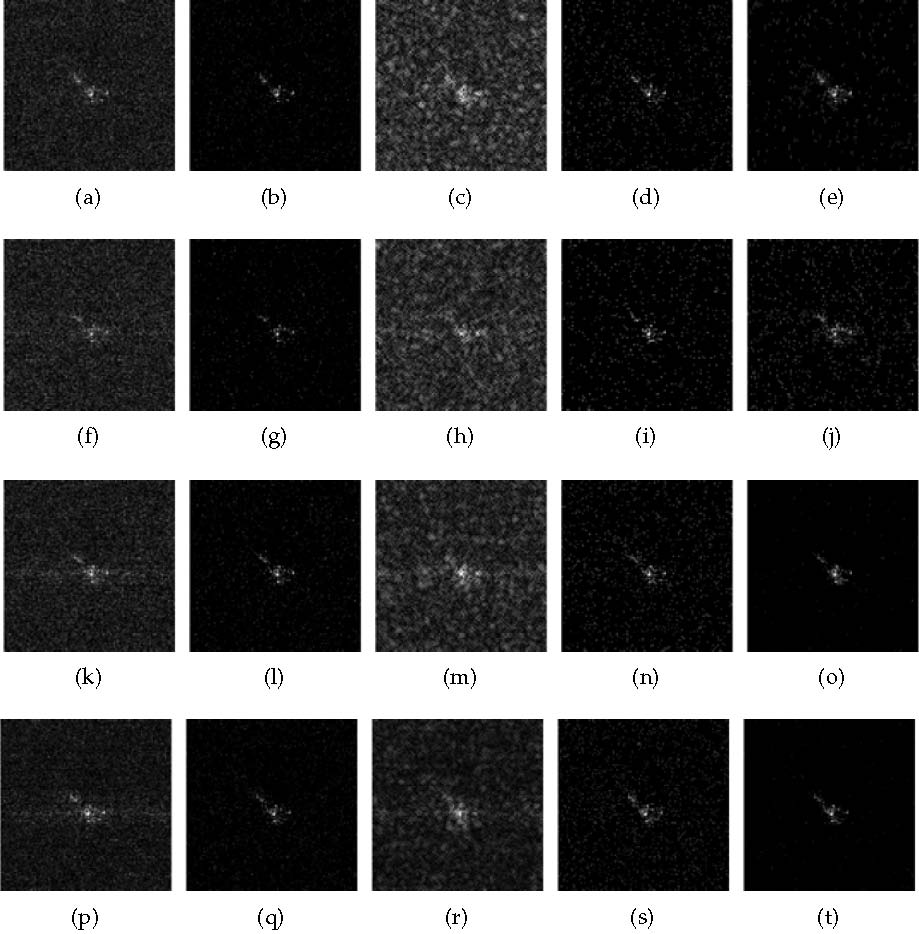}\\
  \caption{Recovered SAR image from undersampled phase history data with MASK-3 according to different imaging algorithms. Row: (a)-(e) $\eta_c=70\%$, $\eta_r=30\%$. (f)-(j) $\eta_c=50\%$, $\eta_r=30\%$. (k)-(o) $\eta_c=50\%$, $\eta_r=50\%$. (p)-(t) $\eta_c=50\%$, $\eta_r=70\%$. Column: (a)-(p) PF-Imaging. (b)-(q) Poi-Imaging. (c)-(r) Reg-Imaging. (d)-(s) IRW-$\ell_1$. (e)-(t) Tar-Imaging.}\label{Fig_CrossandrangeSampling}
\end{figure*}

\begin{table*}\small
\centering
\caption{Numerical Evaluation for SAR image Formation from Undersampled Phase History Data with Mask-2 and Mask-3}\label{Tab:crossrangeandrangeSampling}

\begin{tabular}{|c|c|c|c|c|c|c|c|c|c|}
\hline
\multicolumn{2}{|c|}{ BTR} & \multicolumn{2}{c|}{$\eta_c=20\%,~\eta_r=100\%$} & \multicolumn{2}{c|}{$\eta_c=25\%,~\eta_r=100\%$} & \multicolumn{2}{c|}{$\eta_c=50\%,~\eta_r=100\%$}\tabularnewline
\hline
\multicolumn{2}{|c|}{Algorithm} & PTCR  & $\overline{\chi_t}$ & PTCR    & $\overline{\chi_t}$ & PTCR   & $\overline{\chi_t}$\tabularnewline
\hline
\multicolumn{2}{|c|}{PF-Imaging} & 19.4720    & 18.9968 &  20.1229   & 26.9835   & 22.9048  & 71.6389     \tabularnewline
\hline
\multicolumn{2}{|c|}{Poi-Imaging} & 36.5063   & 49.3023  &36.0164   & 66.4116  & 33.2154   & 126.6901     \tabularnewline
\hline
\multicolumn{2}{|c|}{Reg-Imaging} &   17.2455   & 3.1432   &18.4907   &5.0212   &23.7910   &  17.9493     \tabularnewline
\hline
\multicolumn{2}{|c|}{IRW-$\ell_{1}$} &   31.7427  & \textbf{61.6018}  &31.5526    &  76.8384      &  29.9430    &  132.1517  \tabularnewline
\hline
\multicolumn{2}{|c|}{Tar-Imaging} & \textbf{37.8679}   & 61.0847 & \textbf{51.0485} &\textbf{89.9441}  &  \textbf{67.4151}
   & \textbf{147.0478}  \tabularnewline
\hline
\hline
\multicolumn{2}{|c|}{$\eta_c=70\%,~\eta_r=30\%$} & \multicolumn{2}{c|}{$\eta_c=50\%,~\eta_r=30\%$} & \multicolumn{2}{c|}{$\eta_c=50\%,~\eta_r=50\%$} & \multicolumn{2}{c|}{$\eta_c=50\%,~\eta_r=70\%$}\tabularnewline
\hline
PTCR   & $\overline{\chi_t}$ & PTCR  & $\overline{\chi_t}$ & PTCR  & $\overline{\chi_t}$ & PTCR &$\overline{\chi_t}$ \tabularnewline
\hline
 18.5537  &  14.4999    &  17.2832  & 8.8997     & 19.4325 &  21.5211 & 21.1940   & 38.7339   \tabularnewline
\hline
39.4840 &  126.2416    &  40.8699  & 89.2191    & 37.7610    & 96.6847   & 35.2627    &93.3363   \tabularnewline
\hline
 16.3993  &  2.4573     & 15.0495  & 1.2889   &17.7144     & 4.0257  & 20.4397    &   8.4980    \tabularnewline
\hline
35.5973  &   153.8319    & 36.3232  &  112.3754  &34.0552  & 128.0656   &  32.3910     & 131.9222  \tabularnewline
\hline
\textbf{54.7527} & \textbf{200.9848}   & \textbf{43.0305} & \textbf{148.0628} & \textbf{58.5966} & \textbf{142.3139}  & \textbf{60.6819}   &  \textbf{150.5373}   \tabularnewline
\hline
%
\hline
\multicolumn{2}{|c|}{T72} & \multicolumn{2}{c|}{$\eta_c=20\%,~\eta_r=100\%$} & \multicolumn{2}{c|}{$\eta_c=25\%,~\eta_r=100\%$} & \multicolumn{2}{c|}{$\eta_c=50\%,~\eta_r=100\%$}\tabularnewline
\hline
\multicolumn{2}{|c|}{Algorithm} & PTCR  & $\overline{\chi_t}$ & PTCR    & $\overline{\chi_t}$ & PTCR   & $\overline{\chi_t}$\tabularnewline
\hline
\multicolumn{2}{|c|}{PF-Imaging} &16.5688    &16.4630   & 17.0673 & 22.6121   &17.4500  &  60.1577     \tabularnewline
\hline
\multicolumn{2}{|c|}{Poi-Imaging} & 26.5466   &41.3372  &  26.1173  & 58.6055  &  22.3225  &  109.4114      \tabularnewline
\hline
\multicolumn{2}{|c|}{Reg-Imaging} &14.9497     & 2.9080 &16.3057   & 3.5198 &17.8552    & 14.4514     \tabularnewline
\hline
\multicolumn{2}{|c|}{IRW-$\ell_{1}$} & 23.3785  &  50.1609 &23.1091  &62.3615  &  20.5810    &  111.6243  \tabularnewline
\hline
\multicolumn{2}{|c|}{Tar-Imaging} & \textbf{32.2291}   &\textbf{58.1908}  &\textbf{34.0190}  & \textbf{80.3175}  & \textbf{38.3358}   &  \textbf{133.1198} \tabularnewline
\hline
\hline
\multicolumn{2}{|c|}{$\eta_c=70\%,~\eta_r=30\%$} & \multicolumn{2}{c|}{$\eta_c=50\%,~\eta_r=30\%$} & \multicolumn{2}{c|}{$\eta_c=50\%,~\eta_r=50\%$} & \multicolumn{2}{c|}{$\eta_c=50\%,~\eta_r=70\%$}\tabularnewline
\hline
PTCR   & $\overline{\chi_t}$ & PTCR  & $\overline{\chi_t}$ & PTCR  & $\overline{\chi_t}$ & PTCR &$\overline{\chi_t}$ \tabularnewline
\hline
 13.8400 &  13.1822   &  13.7775  &   8.1656   &  15.0942  &   18.0869   & 15.8337    & 33.4461   \tabularnewline
\hline
  25.8153&  98.4922   & 27.5579 & 71.3681   &  25.1231  &  88.0165  & 23.0272     & 83.8022   \tabularnewline
\hline
12.7821 &  2.2843    &  12.3372 & 1.2096    &   14.2741   &  3.0457   & 15.5342    & 6.6240  \tabularnewline
\hline
 24.1176  & 124.4063    & 25.7013  &  94.5736    & 23.4015  &  109.1500   & 21.7270   & 111.0157   \tabularnewline
\hline
 \textbf{31.3624}& \textbf{153.8931}   & \textbf{31.3459} & \textbf{119.8822} &  \textbf{34.2360} &  \textbf{126.7842} & \textbf{35.4924}   &   \textbf{128.7525}   \tabularnewline
\hline
%
\hline
\multicolumn{2}{|c|}{ BMP2} & \multicolumn{2}{c|}{$\eta_c=20\%,~\eta_r=100\%$} & \multicolumn{2}{c|}{$\eta_c=25\%,~\eta_r=100\%$} & \multicolumn{2}{c|}{$\eta_c=50\%,~\eta_r=100\%$}\tabularnewline
\hline
\multicolumn{2}{|c|}{Algorithm} & PTCR  & $\overline{\chi_t}$ & PTCR   & $\overline{\chi_t}$ & PTCR  & $\overline{\chi_t}$\tabularnewline
\hline
\multicolumn{2}{|c|}{PF-Imaging} & 21.4360    &41.7758   & 22.3384  & 48.9530  &  24.8024  & 132.4351     \tabularnewline
\hline
\multicolumn{2}{|c|}{Poi-Imaging} &  38.3302    &88.9851  &38.2651   &112.0608   & 34.9934     & 215.0542     \tabularnewline
\hline
\multicolumn{2}{|c|}{Reg-Imaging} & 19.5468    &5.8026  & 21.4904   &8.1553   &  27.0500   &34.9255     \tabularnewline
\hline
\multicolumn{2}{|c|}{IRW-$\ell_{1}$} &33.0347    & 99.4547  &33.2900  &117.2942  &  31.5695    &  213.9724  \tabularnewline
\hline
\multicolumn{2}{|c|}{Tar-Imaging} & \textbf{45.5195}   & \textbf{105.2211} &\textbf{58.1724}  &\textbf{141.9737}  & \textbf{68.2833}   & \textbf{236.0894}  \tabularnewline
\hline
\hline
\multicolumn{2}{|c|}{$\eta_c=70\%,~\eta_r=30\%$} & \multicolumn{2}{c|}{$\eta_c=50\%,~\eta_r=30\%$} & \multicolumn{2}{c|}{$\eta_c=50\%,~\eta_r=50\%$} & \multicolumn{2}{c|}{$\eta_c=50\%,~\eta_r=70\%$}\tabularnewline
\hline
PTCR  & $\overline{\chi_t}$ & PTCR  & $\overline{\chi_t}$ & PTCR  & $\overline{\chi_t}$ & PTCR &$\overline{\chi_t}$ \tabularnewline
\hline
 20.0459   &   22.7768   &  18.5443 &    14.6007   &  20.6941  & 36.1805  &  22.8670   & 74.8243   \tabularnewline
\hline
  41.0814 & 154.5493   &  42.6230  &   137.0401   &  39.0203  & 134.0967 &  36.9405    &  148.7263  \tabularnewline
\hline
 18.0004 &    2.9916  &  15.8433 &   1.6735   & 19.2744   & 6.3297   &   23.3238   &  17.0751  \tabularnewline
\hline
37.3317 &  218.1727    &  38.0015 &169.9341   &   35.4457 & 190.5513  &  34.1575    &   214.8892    \tabularnewline
\hline
\textbf{59.2517} &   \textbf{290.5725}  & \textbf{49.0573} & \textbf{245.5196}  & \textbf{59.7272} &  \textbf{213.8323} &    \textbf{62.7116} &   \textbf{229.5607}   \tabularnewline
\hline
\end{tabular}

\end{table*}
\par Intuitively speaking, compared with the Mask-1, the above conclusions also hold that our Tar-Imaging outperforms the other algorithms in the both visual and quantitative results. It is more specially in the results that there will be some fake target points appearing in the cross range in Figs. 13(a),(f) and (k) for PF-Imaging because of the reduced observation angels. However, these fake scattering points are all suppressed in Figs. 13(o) with our framework. If we compare Fig. 14(t) with Fig. 13(o) and Fig. 14(o), we will observe that when $\eta_c=50\%$ and $\eta_r\geq 50$, our framework can produce the similar target enhanced images from visual sight. On the contrary, it can be obviously noticed that the background clutters will get stronger as the range sampling rate $\eta_r$ reducing for other algorithms. We can also obtain this conclusion from Table \ref{Tab:crossrangeandrangeSampling}. Taking BTR as an example, the PTCR for Tar-Imaging is 67.4151(dB) when $\eta_c=50\%,~\eta_r=1$ and 58.5966(dB) when  $\eta_c=50\%,~\eta_r=50\%$ with almost 12(dB) decrease on average. PTCR for IRW-$\ell_1$ instead increases from 29.9430(dB) to 34.0552(dB) when $\eta_r$ is reduced from $1$ to $50\%$. This phenomenon is also easy to understand because an increase of sampling rate will simultaneously enhance all target and background scatters without considering their semantic information. Therefore, the total magnitude of background clutters will become larger, leading to a decreased PTCR in the compared algorithms. This phenomenon further establishes the superiority of our framework by considering the semantic label for each pixel.
\section{Concluding Remarks}\label{Sec:Conclusion}
In this paper, we develop a novel semantic information guided iterative framework for target oriented SAR image formation, which can effectively enhance the target scatters and suppress the background clutters simultaneously. In this framework, two types of semantic information guided regularization functions are developed for the underlying image and its semantic labels, respectively. Compared with the sparsity-regularized imaging algorithms, a plenty of experimental results demonstrate the superiorities of our framework in both visual and quantitative evaluations. The proposed framework sheds a new light on bridging the prior distribution to a simple reweighted $\ell_1$ norm. Based on this idea, much more types of regularizers could be further derived from different distributions for the possible future research. The main insufficiency of our framework is that it can only deal with the simple target scene with a few types of semantic contents. When the scene becomes intricate containing various types of objects, the accuracy of semantic label inference will be degraded so as to weaken the target quality in the generated image.
\par  Some possible directions will be considered as our future researches. Firstly, we will evaluate the ATR performance with our imaging results. In addition to ATR, imaging frameworks driven by other high level perception tasks can be also developed in the future, such as imaging for segmentation. Next, some more informative high level semantic priors can be exploited and involved in the imaging procedure so that the framework will be robust to the intricate environment such as multiple targets. Finally, more representative features can be exploited or directly learned during the imaging process, which will correspondingly improve the imaging quality.

\bibliographystyle{IEEEtran}
\bibliography{IEEEabrv,egbib}

\begin{thebibliography}{10}
\providecommand{\url}[1]{#1}
\csname url@samestyle\endcsname
\providecommand{\newblock}{\relax}
\providecommand{\bibinfo}[2]{#2}
\providecommand{\BIBentrySTDinterwordspacing}{\spaceskip=0pt\relax}
\providecommand{\BIBentryALTinterwordstretchfactor}{4}
\providecommand{\BIBentryALTinterwordspacing}{\spaceskip=\fontdimen2\font plus
\BIBentryALTinterwordstretchfactor\fontdimen3\font minus
  \fontdimen4\font\relax}
\providecommand{\BIBforeignlanguage}[2]{{%
\expandafter\ifx\csname l@#1\endcsname\relax
\typeout{** WARNING: IEEEtran.bst: No hyphenation pattern has been}%
\typeout{** loaded for the language `#1'. Using the pattern for}%
\typeout{** the default language instead.}%
\else
\language=\csname l@#1\endcsname
\fi
#2}}
\providecommand{\BIBdecl}{\relax}
\BIBdecl

\bibitem{Hummel2000}
R.~Hummel, ``Model-based atr using synthetic aperture radar,'' in \emph{IEEE
  Int. Conf. Radar}.\hskip 1em plus 0.5em minus 0.4em\relax IEEE, 2000, pp.
  856--861.

\bibitem{Carrara1995}
W.~G. Carrara, R.~S. Goodman, and R.~M. Majewski, \emph{Spotlight synthetic
  aperture radar- Signal processing algorithms(Book)}, 1995.

\bibitem{Desai1992Convolution}
M.~D. Desai and W.~K. Jenkins, ``Convolution backprojection image
  reconstruction for spotlight mode synthetic aperture radar,'' \emph{{IEEE}
  Trans. Image Process.}, vol.~1, no.~4, pp. 505--517, Apr. 1992.

\bibitem{Cetin2014}
M.~$\c{C}$etin, I.~Stojanovi$\acute{c}$, N.~$\ddot{O}$nhon, K.~Varshney,
  S.~Samadi, W.~Karl, and A.~Willsky, ``Sparsity-driven synthetic aperture
  radar imaging: Reconstruction, autofocusing, moving targets, and compressed
  sensing,'' \emph{{IEEE} Signal Process. Mag.}, vol.~31, no.~4, pp. 27--40,
  Jul. 2014.

\bibitem{Zhang2014Attributed}
Y.~Zhang, C.~He, X.~Xu, and M.~Liao, ``Attributed scattering center feature
  extraction of high resolution sar image and classification algorithm,'' in
  \emph{IEEE Int. Geosci. Remote Sens. Symp. (IGARSS)}, 2014, pp. 474--477.

\bibitem{cetin2001feature}
M.~\c{C}etin and W.~C. Karl, ``Feature-enhanced synthetic aperture radar image
  formation based on nonquadratic regularization,'' \emph{{IEEE} Trans. Image
  Process.}, vol.~10, no.~4, pp. 623--631, Apr. 2001.

\bibitem{franklin1974tikhonov}
J.~N. Franklin, ``On tikhonov's method for ill-posed problems,'' \emph{Math.
  Comput.}, vol.~28, no. 128, pp. 889--907, 1974.

\bibitem{Osher2005}
S.~Osher, M.~Burger, D.~Goldfarb, J.~Xu, and W.~Yin, ``An iterative
  regularization method for total variation-based image restoration,''
  \emph{Multiscale Modeling and Simulation}, vol.~4, no.~2, pp. 460--489, 2005.

\bibitem{cetin2003feature}
M.~\c{C}etin, W.~C. Karl, D.~Casta{\~n}on \emph{et~al.}, ``Feature enhancement
  and atr performance using nonquadratic optimization-based sar imaging,''
  \emph{{IEEE} Trans. Aerosp. Electron. Syst.}, vol.~39, no.~4, pp. 1375--1395,
  Apr. 2003.

\bibitem{Kelly2012}
S.~I. Kelly, C.~Du, G.~Rilling, and M.~E. Davies, ``Advanced image formation
  and processing of partial synthetic aperture radar data,'' \emph{IET Signal
  Process.}, vol.~6, no.~5, pp. 511--520, 2012.

\bibitem{baraniuk2007compressive}
R.~G. Baraniuk, ``Compressive sensing,'' \emph{{IEEE} Signal Process. Mag.},
  vol.~24, no.~4, Jul. 2007.

\bibitem{donoho2006compressed}
D.~L. Donoho, ``Compressed sensing,'' \emph{{IEEE} Trans. Inf. Theory},
  vol.~52, no.~4, pp. 1289--1306, Apr. 2006.

\bibitem{candes2006robust}
E.~J. Cand{\`e}s, J.~Romberg, and T.~Tao, ``Robust uncertainty principles:
  Exact signal reconstruction from highly incomplete frequency information,''
  \emph{{IEEE} Trans. Inf. Theory}, vol.~52, no.~2, pp. 489--509, Feb. 2006.

\bibitem{Candes2011}
E.~J. Cand{\`e}s and Y.~Plan, ``Tight oracle inequalities for low-rank matrix
  recovery from a minimal number of noisy random measurements,'' \emph{{IEEE}
  Trans. Inf. Theory}, vol.~57, no.~4, pp. 2342--2359, Apr. 2011.

\bibitem{TelloAlonso2010}
M.~Tello~Alonso, P.~Lopez-Dekker, and J.~Mallorqui, ``A novel strategy for
  radar imaging based on compressive sensing,'' \emph{{IEEE} Trans. Geosci.
  Remote Sens.}, vol.~48, no.~12, pp. 4285--4295, Dec. 2010.

\bibitem{Patel2010}
V.~Patel, G.~Easley, J.~Healy, D.M., and R.~Chellappa, ``Compressed synthetic
  aperture radar,'' \emph{{IEEE} J. Sel. Topics Signal Process.}, vol.~4,
  no.~2, pp. 244--254, Apr. 2010.

\bibitem{Samadi2011}
S.~Samadi, M.~\c{C}etin, and M.~Masnadi-shirazi, ``Sparse representation-based
  synthetic aperture radar imaging,'' \emph{IET Radar, Sonar and Navigation},
  vol.~5, no.~2, pp. 182--193, 2011.

\bibitem{Herman2008}
M.~Herman and T.~Strohmer, ``Compressed sensing radar,'' in \emph{IEEE Radar
  Conf.}\hskip 1em plus 0.5em minus 0.4em\relax IEEE, 2008, pp. 1--6.

\bibitem{Herman2009}
M.~A. Herman and T.~Strohmer, ``High-resolution radar via compressed sensing,''
  \emph{{IEEE} Trans. Signal Process.}, vol.~57, no.~6, pp. 2275--2284, Jun.
  2009.

\bibitem{Baraniuk2007}
R.~Baraniuk and P.~Steeghs, ``Compressive radar imaging,'' in \emph{IEEE Radar
  Conf.}\hskip 1em plus 0.5em minus 0.4em\relax IEEE, 2007, pp. 128--133.

\bibitem{Zhang2010}
L.~Zhang, M.~Xing, C.-W. Qiu, J.~Li, J.~Sheng, Y.~Li, and Z.~Bao, ``Resolution
  enhancement for inversed synthetic aperture radar imaging under low snr via
  improved compressive sensing,'' \emph{{IEEE} Trans. Geosci. Remote Sens.},
  vol.~48, no.~10, pp. 3824--3838, Oct. 2010.

\bibitem{Zhao2016}
L.~Zhao, L.~Wang, G.~Bi, S.~Li, L.~Yang, and H.~Zhang, ``Structured
  sparsity-driven autofocus algorithm for high-resolution radar imagery,''
  \emph{Signal Process.}, vol. 125, pp. 376 -- 388, 2016.

\bibitem{Potter2010}
L.~Potter, E.~Ertin, J.~Parker, and M.~Cetin, ``Sparsity and compressed sensing
  in radar imaging,'' \emph{Proc. {IEEE}}, vol.~98, no.~6, pp. 1006--1020, Jun.
  2010.

\bibitem{Smith1997}
S.~M. Smith and J.~M. Brady, ``Susan¡ªa new approach to low level image
  processing,'' \emph{Int. J. Comput. Vision}, vol.~23, no.~1, pp. 45--78,
  1997.

\bibitem{wang2014enhanced}
L.~Wang, L.~Zhao, G.~Bi, C.~Wan, and L.~Yang, ``Enhanced isar imaging by
  exploiting the continuity of the target scene,'' \emph{{IEEE} Trans. Geosci.
  Remote Sens.}, vol.~52, no.~9, pp. 5736--5750, 2014.

\bibitem{Ji2008}
S.~Ji, Y.~Xue, and L.~Carin, ``Bayesian compressive sensing,'' \emph{{IEEE}
  Trans. Signal Process.}, vol.~56, no.~6, pp. 2346--2356, 2008.

\bibitem{Wang2015}
L.~Wang, L.~Zhao, G.~Bi, and C.~Wan, ``Sparse representation-based isar imaging
  using markov random fields,'' \emph{IEEE J. Sel. Topics Appl. Earth
  Observations Remote Sens.}, vol.~8, no.~8, pp. 3941--3953, Aug. 2015.

\bibitem{Feng2016}
L.~Feng and B.~Bhanu, ``Semantic concept co-occurrence patterns for image
  annotation and retrieval,'' \emph{{IEEE} Trans. Pattern Anal. Mach. Intell.},
  vol.~38, no.~4, pp. 785--799, Apr. 2016.

\bibitem{Bengio2009a}
Y.~Bengio, ``Learning deep architectures for ai,'' \emph{Found. Trends in
  Machine Learning}, vol.~2, no.~1, pp. 1--127, 2009.

\bibitem{Wen2013}
Z.~Wen, B.~Hou, and S.~Wang, ``High resolution sar target reconstruction from
  compressive measurements with prior knowledge,'' in \emph{IEEE Int. Geosci.
  Remote Sens. Symp. (IGARSS), 2013}, Jul. 2013, pp. 3167--3170.

\bibitem{Shannon1949}
C.~E. Shannon, ``Communication in the presence of noise,'' \emph{Proc. IRE},
  vol.~37, no.~1, pp. 10--21, 1949.

\bibitem{Gui2010}
G.~Gui, ``Statistical modeling of sar images: A survey,'' \emph{Sensors},
  vol.~10, no.~1, pp. 775--795, 2010.

\bibitem{Zhang2015}
P.~Zhang, M.~Li, Y.~Wu, and H.~Li, ``Hierarchical conditional random fields
  model for semisupervised sar image segmentation,'' \emph{{IEEE} Trans.
  Geosci. Remote Sens.}, vol.~53, no.~9, pp. 4933--4951, Sept. 2015.

\bibitem{Weisenseel1999}
R.~A. Weisenseel, W.~C. Karl, D.~A. Castanon, G.~J. Power, and P.~Douville,
  ``Markov random field segmentation methods for sar target chips,''
  \emph{Proc. SPIE}, vol. 3721, pp. 462--473, 1999.

\bibitem{feng2013multiphase}
J.~Feng, Z.~Cao, and Y.~Pi, ``Multiphase sar image segmentation
  with-{${G}^0$}statistical-model-based active contours,'' \emph{{IEEE} Trans.
  Geosci. Remote Sens.}, vol.~51, no.~7, pp. 4190--4199, Jul. 2013.

\bibitem{Krylov2011}
V.~A. Krylov, G.~Moser, S.~B. Serpico, and J.~Zerubia, ``Supervised
  high-resolution dual-polarization sar image classification by finite mixtures
  and copulas,'' \emph{{IEEE} J. Sel. Topics Signal Process.}, vol.~5, no.~3,
  pp. 554--566, Mar. 2011.

\bibitem{Ghinelli1997The}
B.~M.~G. Ghinelli and J.~C. Bennett, ``The application of artificial neural
  networks and standard statistical methods to sar image classification,'' in
  \emph{IEEE Int. Geosci. Remote Sens. Symp. (IGARSS), 1997}, 1997, pp.
  1211--1213 vol.3.

\bibitem{Zhang2008}
X.~Zhang, L.~Jiao, F.~Liu, L.~Bo, and M.~Gong, ``Spectral clustering ensemble
  applied to sar image segmentation,'' \emph{{IEEE} Trans. Geosci. Remote
  Sens.}, vol.~46, no.~7, pp. 2126--2136, Jul. 2008.

\bibitem{Ainsworth2008}
T.~L. Ainsworth, D.~L. Schuler, and J.~S. Lee, ``Polarimetric sar
  characterization of man-made structures in urban areas using normalized
  circular-pol correlation coefficients,'' \emph{Remote Sensing of
  Environment}, vol. 112, no.~6, pp. 2876--2885, 2008.

\bibitem{Li2009}
S.~Z. Li, \emph{Markov random field modeling in image analysis}.\hskip 1em plus
  0.5em minus 0.4em\relax Springer Science \& Business Media, 2009.

\bibitem{Boykov2001Fast}
Y.~Boykov, O.~Veksler, and R.~Zabih, ``Fast approximate energy minimization via
  graph cuts,'' \emph{{IEEE} Trans. Pattern Anal. Mach. Intell.}, vol.~1,
  no.~11, pp. 1222--1239, 2001.

\bibitem{Tao1998}
P.~D. Tao and T.~H. An, ``A d.c. optimization algorithm for solving the
  trust-region subproblem,'' \emph{SIAM J. Opt.}, vol.~8, no.~2, pp. 476--505,
  1998.

\bibitem{Schuele2005}
T.~Sch$\ddot{u}$le, C.~Schn$\ddot{o}$rr, S.~Weber, and J.~Hornegger, ``Discrete
  tomography by convex$-$concave regularization and d.c. programming,''
  \emph{Discrete Appl. Math.}, vol. 151, no. 1-3, pp. 229--243, 2005.

\bibitem{beck2009fast}
A.~Beck and M.~Teboulle, ``A fast iterative shrinkage-thresholding algorithm
  for linear inverse problems,'' \emph{SIAM J. Imag. Sci.}, vol.~2, no.~1, pp.
  183--202, Jan. 2009.

\bibitem{Becker2011}
S.~B$\acute{e}$cker, J.~Bobin, and E.~J. Cand$\grave{e}$s, ``Nesta: A fast and
  accurate first-order method for sparse recovery,'' \emph{SIAM J. Imag. Sci.},
  vol.~4, no.~1, pp. 1--39, Jan. 2011.

\bibitem{Nesterov1983}
Y.~E. Nesterov, ``A method for solving the convex programming problem with
  convergence rate $o(1/ksp{2})$.'' \emph{Dokl.akad.nauk Sssr}, no.~3, pp.
  543--547, 1983.

\bibitem{Besag1986}
J.~Besag, ``On the statistical-analysis of dirty pictures,'' \emph{J. Roy.
  Stat. Soc.}, vol. b-48, no.~3, pp. 259--302, 1986.

\bibitem{Bezdek1984}
J.~C. Bezdek, R.~Ehrlich, and W.~Full, ``Fcm: The fuzzy c-means clustering
  algorithm,'' \emph{Comput. Geosci.}, vol.~10, no.~84, pp. 191--203, 1984.

\bibitem{Candes2008}
E.~J. Cand$\grave{e}$s, M.~B. Wakin, and S.~P. Boyd, ``Enhancing sparsity by
  reweighted $\ell$1 minimization,'' \emph{J. Fourier Anal. Applicat.},
  vol.~14, no. 5-6, pp. 877--905, 2008.

\bibitem{Ma2015}
P.~Ma, H.~Lin, H.~Lan, and F.~Chen, ``On the performance of reweighted $l_{1}$
  minimization for tomographic sar imaging,'' \emph{{IEEE} Geosci. Remote Sens.
  Lett.}, vol.~12, no.~4, pp. 895--899, Apr. 2015.

\end{thebibliography}

%

%
%
%




\end{document}